\documentclass[10pt,twocolumn,letterpaper]{article}

\usepackage[pagenumbers]{cvpr} %

\usepackage{cuted} %

\usepackage{currfile} %

\usepackage{caption} %

\usepackage{array}
\usepackage[table]{xcolor}
\usepackage{adjustbox}
\usepackage{multirow}

\definecolor{cvprblue}{rgb}{0.21,0.49,0.74}
\definecolor{s1band}{RGB}{240,240,240}
\definecolor{s1overall}{RGB}{239,239,239}
\definecolor{s1blueA}{RGB}{143,184,236}
\definecolor{s1blueB}{RGB}{188,213,244}
\definecolor{s1blueC}{RGB}{226,236,250}
\definecolor{s1greenA}{RGB}{140,211,192}
\definecolor{s1greenB}{RGB}{191,231,220}
\definecolor{s1greenC}{RGB}{224,243,238}
\definecolor{s1gain}{RGB}{0,158,115}
\definecolor{s1loss}{RGB}{213,94,0}
\definecolor{reqbg}{RGB}{245,247,250}
\definecolor{reqborder}{RGB}{208,215,226}
\definecolor{jsonkey}{RGB}{29,95,168}
\definecolor{jsonstr}{RGB}{15,118,110}
\definecolor{jsonnum}{RGB}{194,65,12}
\definecolor{goldmark}{RGB}{255,229,138}
\definecolor{notegrey}{RGB}{107,114,128}
\usepackage[pagebackref,breaklinks,colorlinks,allcolors=cvprblue]{hyperref}

\def\confName{CVPR}
\def\confYear{2027}

\newcommand{\nQuestions}{17,500}
\newcommand{\nTasks}{35}

\newcommand{\nBenchTasks}{5}

\newcommand{\nShortQuestions}{11,500}
\newcommand{\nLongQuestions}{6,000}
\newcommand{\nAVQuestions}{1,500}
\newcommand{\nAVTasks}{3}
\newcommand{\nShortTasks}{20}
\newcommand{\nLongTasks}{7}
\newcommand{\nStaticTasks}{32}

\newcommand{\nEnvs}{102}
\newcommand{\nPlayableEnvs}{97}
\newcommand{\nEnvSuites}{3}

\newcommand{\nModels}{33}
\newcommand{\nSystemOne}{12}
\newcommand{\nOpenMLLMs}{21}

\newcommand{\nAudQuestions}{10,560}
\newcommand{\nAudTasks}{19}
\newcommand{\nAudSources}{21}
\newcommand{\nAudModels}{14}
\newcommand{\nImgQuestions}{23,000}
\newcommand{\nImgTasks}{40}
\newcommand{\nImgSources}{47}
\newcommand{\nImgStd}{7}
\newcommand{\nImgStdQuestions}{3,454}

\newcommand{\nAllQuestions}{51,060}
\newcommand{\nAllTasks}{94}

\newcommand{\nNoAudModels}{19}
\newcommand{\nNoAudCoverage}{76.4}
\newcommand{\nStreamEpisodes}{631}
\newcommand{\nAllDomains}{16}

\newcommand{\nThinkModels}{13}
\newcommand{\nNoThinkModels}{8}
\newcommand{\nRecipeSourcesRelease}{10}
\newcommand{\nPublicSources}{94}

\newcommand{\name}{S1Arena-MM}

\title{\name: Which Multimodal Decisions Can Be Left to System 1?}

\author{
\bfseries Lang Lin$^{1}$\thanks{Equal contribution.} \quad Tianyu Chen$^{1}$\footnotemark[1] \quad
Yi Gu$^{1}$\footnotemark[1] \quad Yang Zhao$^{1}$\footnotemark[1] \quad
Xinran Song$^{1}$\footnotemark[1] \quad Xinyang Liu$^{1}$\footnotemark[1]\\
\bfseries Jianpeng Cheng$^{2}$ \quad Mingyuan Zhou$^{1}$\thanks{Corresponding author: \texttt{mingyuan.zhou@mccombs.utexas.edu}}\\
$^{1}$The University of Texas at Austin \qquad $^{2}$Meta AI
}

\begin{document}
\maketitle

\begin{abstract}
Perception systems often make discrete decisions: flag a hateful meme, predict whether a pedestrian will cross, detect a
spoofed voice, or select the next move in a game. \emph{System-1 models}, such as OneJev and Jev-Omni, read the
media and return probabilities over the options in a single forward pass, while multimodal large language models
(MLLMs) generate text. Which multimodal decisions can System-1 models handle, and where do MLLMs perform better? We
introduce \name{}, a benchmark that gives every model the same decision requests and scores the probabilities it
returns. It contains \nAllQuestions{} questions across \nAllTasks{} tasks with image, video, and audio input, drawn
from \nPublicSources{} public datasets and three program-generated ones.
It also includes \nEnvs{} closed-loop game environments
that a model plays from screenshots; in real-time mode, the game continues while the model decides. We evaluate
\nModels{} open models: \nSystemOne{} System-1 models and \nOpenMLLMs{} MLLMs read at the answer token. On static
decisions, the best System-1 model leads the best MLLM by 4.9 percentage points on images and is within one point of it
on video and audio. Post-training for decisions mainly improves the probabilities: on video, a System-1 model has a
lower Brier score than the MLLM that it is post-trained from in 10 of 12 pairs. With up to 2{,}048 thought tokens per
question, the most accurate MLLMs pass the best System-1 models only slightly, and thinking raises the Brier score of
most MLLMs.
Closed loops separate the models: three System-1 models built on a 27B MLLM reach 22.0--33.3\% progression in BALROG,
and no other model exceeds 9.7\% with thinking off. In real-time GameWorld, latencies of up to 1.1~s per decision cause
no measurable loss of progress for System-1 models.
Code and data: \mbox{\href{https://github.com/zhaoyang02/S1Arena-MM}{\nolinkurl{github.com/zhaoyang02/S1Arena-MM}}}.
\end{abstract}

\section{Introduction}
\label{sec:intro}

\begin{figure*}[t]
\centering
\includegraphics[width=\textwidth]{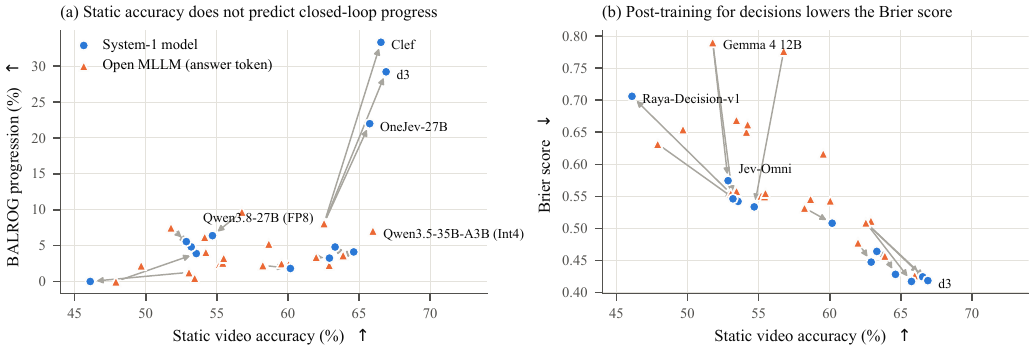}
\caption{\textbf{Static accuracy does not predict closed-loop progress, and post-training for decisions mainly improves
the probabilities.} One point per model; a gray arrow goes from each base MLLM to the System-1 model post-trained from
it. Static: the \nStaticTasks{} video tasks that every model takes. (a) The four models with the highest static
accuracy (65.8--66.9\%) reach 7.0--33.3\% progression in BALROG. (b) In 10 of 12 pairs, the System-1 model has a lower
Brier score than its base, while its accuracy changes by $-6.9$ to $+5.6$ points.}
\label{fig:front}
\end{figure*}

Many perception applications make discrete decisions.
A moderation system flags a hateful meme, a driving system predicts whether a pedestrian will cross, a voice login
detects a spoofed voice, and a game agent selects its next move.
These requests occur at high volume, and each needs an option with a probability.
Some also occur under a time limit: while the model decides, the scene continues to change.
Multimodal large language models (MLLMs) can decide by generating text, but generation adds latency and cost, and the
answer gives no reliable probability.

\emph{System~1} makes fast, automatic judgments, while \emph{System~2} reasons slowly and deliberately
\cite{kahneman2011thinking}.
Here, we study multimodal \emph{System-1 models}.
OneJev \cite{omnijev2026onejev}, Jev-Omni \cite{jevomni2026}, Clef \cite{cloudflare2026clef},
Raya \cite{raya2026decision}, and Xor \cite{juspay2026xor} accept a state with images, video, or audio, and typed
questions.
For each question, they return a probability distribution in one forward pass and generate no text.
Zero-shot classifiers such as CLIP \cite{radford2021clip} also score options in one pass, but they take only label
texts, without an instruction or a question. System-1 models take instructions and option descriptions in natural
language.
Each model also sets how it reads the media: the image resolution, the frame rate, the number of frames, and the audio
length.
This reading policy controls both the cost of a decision and the evidence that the model sees.

This raises a central question: which multimodal decisions can System-1 models handle, and where do MLLMs perform
better?
Benchmarks for MLLMs, such as Video-MME \cite{fu2025videomme}, MVBench \cite{li2024mvbench}, and EgoSchema
\cite{mangalam2023egoschema}, measure the accuracy of generated answers on question answering.
They do not score probabilities, and they rarely include decisions such as moderation, quality control, judging, and
routing.
Game benchmarks such as BALROG \cite{paglieri2025balrog} and VS-Bench \cite{xu2025vsbench} pause the game while the
model decides, so latency has no effect on the outcome.
The text benchmark System1-Arena~\cite{chen2026system1arena} compares System-1 models and LLMs on the same decisions, but for text only.

We introduce \name{}, a benchmark with \nAllQuestions{} typed questions across \nAllTasks{} tasks with image, video,
and audio input, drawn from \nPublicSources{} public datasets and three program-generated ones.
It also includes \nEnvs{} closed-loop game environments from \nEnvSuites{} suites, which a model plays from
screenshots.
\name{} adopts the decision requests and the metrics of System1-Arena~\cite{chen2026system1arena} and extends them to
images, video, audio, and multimodal closed-loop games.
\name{} follows three design choices.
\begin{itemize}[leftmargin=*,itemsep=0pt,topsep=0pt]
  \item \textbf{Same request, own reading.}
  Every model receives the same request: a state that refers to media, and typed questions.
  Each model reads the media with its own policy, and we report it as part of the model.
  A model that cannot take a modality is not penalized on accuracy; we report its coverage instead
  (Sections~\ref{sec:interface} and~\ref{sec:setup}).
  \item \textbf{More than accuracy.}
  For every question, each model returns a probability for each option.
  We report accuracy, the Brier score, coverage, and latency (Section~\ref{sec:metrics}).
  \item \textbf{Decisions in closed loops.}
  A model plays games from screenshots by choosing legal actions.
  In real-time mode, the game continues during each model call (Section~\ref{sec:loop}).
\end{itemize}

We evaluate \nModels{} open models: \nSystemOne{} System-1 models and \nOpenMLLMs{} MLLMs, which we read at the answer
token, so that every model decides in one forward pass (Figure~\ref{fig:front}).
On static decisions, System-1 models lead on images and are close to the MLLMs on video and audio: the best System-1
model leads the best MLLM by 4.9 percentage points on images and by less than one point on video and audio.
Post-training for decisions mainly improves the probabilities: on video, a System-1 model has a lower Brier score than
the MLLM that it is post-trained from in 10 of 12 pairs, while its accuracy moves in both directions
(Table~\ref{tab:pairs}).
The reading policy is part of the decision: on a needle task that asks whether and where a short clip was inserted into
a long video, OneJev-27B, which reads one frame per second, answers 94.2\% of the questions correctly, and Jev-Omni,
which reads 16 frames, 42.2\%.
Physical outcomes remain hard for every model: none exceeds 61\% on object contact and collisions, where chance is 50\%.

With thinking, the strongest MLLMs pass the best System-1 models, but only slightly and with worse probabilities.
With about 400 thought tokens per question, Qwen3.8-27B passes d3 on images, and with about 1{,}200, Qwen3.5-35B-A3B
passes it on video, but thinking raises the Brier score of most MLLMs (Figure~\ref{fig:thinking}).

Closed loops separate the models more than static decisions do.
The three System-1 models built on Qwen3.8-27B reach 22.0--33.3\% progression in BALROG, while no other model exceeds
9.7\% with thinking off.
In real-time GameWorld, latencies of up to 1.1~s per decision cause no measurable loss of progress.
In spoken conversations, System-1 models stop for most backchannels, although Clef-Omni tells them apart from
interruptions offline.

In summary, we make four contributions.
First, \name{} offers System-1 models and MLLMs a common decision interface over images, video, and audio, with each
label and instruction linked to its source.
Second, we treat the reading policy as part of the model and report coverage, so models with different modalities and
frame budgets can be compared.
Third, a game bridge lets any decision model play \nEnvs{} game environments from screenshots, in paused and real-time
mode. 
Fourth, our evaluation of \nModels{} open models, with paired comparisons of System-1 models and the MLLMs that they are
post-trained from, identifies which multimodal decisions System-1 models can handle.

\section{Related Work}
\label{sec:related}

\paragraph{Video and multimodal benchmarks.}
Video-MME \cite{fu2025videomme}, MVBench \cite{li2024mvbench}, TempCompass \cite{liu2024tempcompass}, NExT-QA
\cite{xiao2021nextqa}, Perception Test \cite{patraucean2023perception}, EgoSchema \cite{mangalam2023egoschema}, and
VSI-Bench \cite{yang2025thinking} evaluate the perception, temporal reasoning, and spatial memory of MLLMs through
question answering.
They score the generated answer.
\name{} includes subsets of five of these benchmarks with their original questions, so that results can be compared with
published numbers.
Its other tasks cover application decisions, such as moderation, quality control, judging generated videos, and model
routing, and it scores a probability for every option.

\paragraph{System~1 and typed decision models.}
The dual-process distinction of \citet{kahneman2011thinking} often describes how much a language model reasons before it
answers \cite{yu2024distilling,ziabari2025spectrum,zhang2025s1bench}.
Zero-shot classifiers such as CLIP \cite{radford2021clip} score label texts in one pass, without an instruction or a
question.
System-1 models, such as the OpenAI Decisions API \cite{openai2026decisions} and Jev \cite{typesafe2026jev}, take
typed questions with options in natural language and return a probability for each option.
System1-Arena~\cite{chen2026system1arena} compares them with LLMs on text decisions.
\name{} extends this comparison to System-1 models that read images, video, and audio.

\paragraph{Calibration and cost.}
The Brier score \cite{brier1950verification} evaluates probabilistic predictions, and the expected calibration error
\cite{guo2017calibration} measures how far confidence is from accuracy.
Cost-aware methods route requests between models of different sizes \cite{chen2023frugalgpt}, and RouterBench
\cite{hu2024routerbench} evaluates such routers.
For video, the number of frames that a model reads sets most of its cost.
We therefore report latency and coverage with each model's reading policy, and include a routing task that asks when to
escalate from a fast model to a slower one.

\paragraph{Games and closed loops.}
BALROG \cite{paglieri2025balrog} evaluates LLMs and vision-language models on BabyAI, Crafter, MiniHack, NetHack, and other
games, and VS-Bench \cite{xu2025vsbench} evaluates vision-language models in multi-agent environments.
GameWorld \cite{ouyang2026gameworld} provides browser games with tasks, and its variant GameWorld-RT keeps the game
running while the model computes.
BALROG and VS-Bench pause the game while the model decides.
\citet{kang2025winfast} choose the model size and the quantization level from the latency budget and do better in a
trading simulation and a fighting game.
We use these suites through one bridge, present each step as a typed decision, and add a real-time mode that charges
each decision the model's latency measured on fixed hardware, so a score does not depend on the serving machine.

\section{\name{}}
\label{sec:benchmark}

\subsection{Decision requests with media}
\label{sec:interface}

\begin{figure}[t]
\centering
\newcommand{\jk}[1]{\textcolor{jsonkey}{"#1"}}%
\newcommand{\js}[1]{\textcolor{jsonstr}{"#1"}}%
\newcommand{\jn}[1]{\textcolor{jsonnum}{#1}}%
\newcommand{\jgold}[1]{{\setlength{\fboxsep}{0.6pt}\colorbox{goldmark}{\textcolor{jsonkey}{"#1"}}}}%
\newcommand{\ind}[1]{\hspace*{#1\fontdimen2\font}}%
\newcommand{\reqhead}[2]{\noindent\begin{minipage}{\columnwidth}\textbf{#1}\hfill
  {\footnotesize\textcolor{notegrey}{#2}}\end{minipage}}%
\newcommand{\reqbox}[1]{\fcolorbox{reqborder}{reqbg}{\parbox{\dimexpr\columnwidth-2\fboxsep-2\fboxrule\relax}{%
  \frenchspacing#1}}}%
\newcommand{\reqframe}[1]{\includegraphics[width=0.235\columnwidth]{figures/request/#1}}%
\newcommand{\reqtime}[1]{{\tiny\textcolor{notegrey}{#1}}}%
\reqhead{Decision request}{a state with a video and a typed question}\\[1pt]
\reqbox{\centering\scriptsize\setlength{\tabcolsep}{0pt}\renewcommand{\arraystretch}{0.9}%
\begin{tabular}{@{}c@{\hspace{1.5pt}}c@{\hspace{1.5pt}}c@{\hspace{1.5pt}}c@{}}
\reqframe{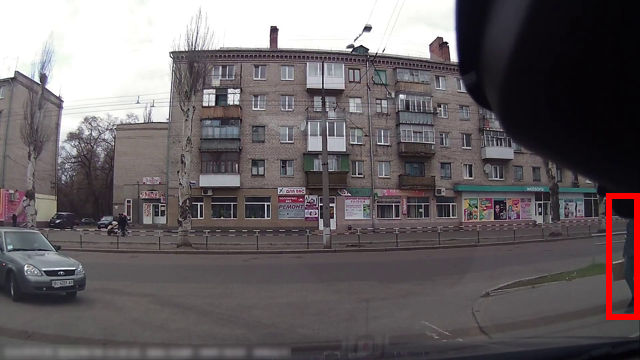} & \reqframe{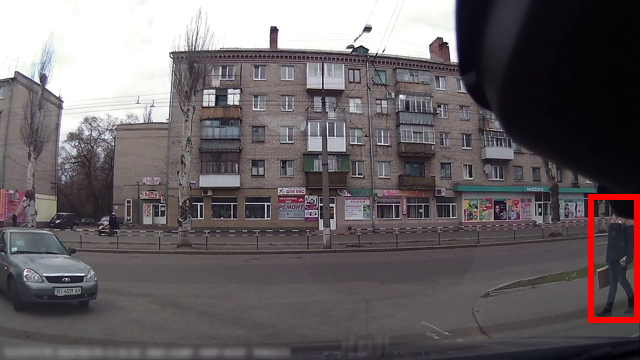} & \reqframe{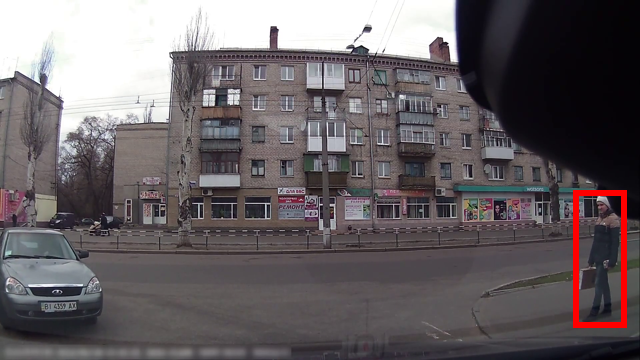} & \reqframe{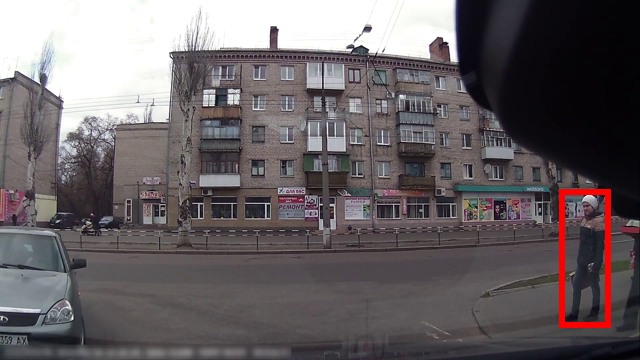}
\\[-1.5pt]
\reqtime{0.3\,s} & \reqtime{0.9\,s} & \reqtime{1.6\,s} & \reqtime{2.2\,s}
\end{tabular}\par\vspace{2pt}\raggedright\ttfamily
\{\\
\ind{2}\jk{state}: \{\\
\ind{4}\jk{text}: \js{Observe the complete video history\\
\ind{6}strictly before the prediction cutoff. ...},\\
\ind{4}\jk{video}: \{\jk{uri}: \js{jaad/videos/video\_0297.mp4},\\
\ind{6}\jk{window}: [\jn{0.0}, \jn{2.5}], \jk{marks}: \{...\}\}\\
\ind{2}\},\\
\ind{2}\jk{questions}: \{\\
\ind{4}\jk{answer}: \{\jk{type}: \js{boolean},\\
\ind{6}\jk{instructions}: \js{Will the pedestrian outlined\\
\ind{8}in red start crossing the road within the\\
\ind{8}next two seconds?}\}\\
\ind{2}\}\\
\}}\\[4pt]
\reqhead{Answers}{d3 and its base MLLM, gold option marked}\\[1pt]
\reqbox{\raggedright\scriptsize\ttfamily
\{\\
\ind{2}\jk{d3}:\ind{10}\{\jgold{true}: \jn{0.85}, \jk{false}: \jn{0.15}\},\\
\ind{2}\jk{Qwen3.8-27B}: \{\jgold{true}: \jn{0.27}, \jk{false}: \jn{0.73}\}\\
\}}
\caption{\textbf{A decision request and the answers of a System-1 model and its base MLLM.} The request is from the
pedestrian-crossing task, the System-1 model is d3, and its base MLLM is Qwen3.8-27B. The request names the video and
the window that the model may see, 0--2.5~s of a 13-s clip, and adds a
mark that outlines the target pedestrian in red. We show four frames spread evenly over the window, draw the outline
thicker than in the input so that it stays visible at this size, and shorten the state text. An open MLLM gets the same
request as a prompt (Appendix~\ref{app:prompt}).}
\label{fig:request}
\end{figure}

Following System1-Arena \cite{chen2026system1arena}, we represent every static example and game step as a
\emph{decision request} $r = (s, Q)$ (Figure~\ref{fig:request}).
The state $s$ contains named fields.
A field holds text, a reference to an image, a video, or an audio track, or an ordered list of references, such as the
views of several cameras.
A reference names the file, its hash, and a time window; it can also add marks, such as a box around the target
pedestrian, which are drawn onto the frames.
A text field can place the $k$-th image of the state at a marker, so a game step can interleave screenshots with the
history.
The questions $Q = \{q_1, \dots, q_m\}$ share this state.

A question is \textbf{boolean} (yes/no), \textbf{choice} (2--255 options, exactly one of them correct), or
\textbf{score} (2--10 ordered levels).
Each question comes with instructions and each option with a description, so no model needs a dataset-specific
prompt.
For each question, the model outputs a probability distribution over its options.
The reference is a hard label or, if the source provides one, a distribution of labels.

\paragraph{Reading policy.}
The request stores the original media and the window, not frames sampled from the video or audio resampled to a fixed
rate.
Each model reads the media with its own policy: the frame rate, the maximum number of frames, the resolution, and the
audio rate.
The policy is part of the model's identity.
A request carries exactly the media kinds that it needs.
If a model does not take one of them, the runner makes no call and marks the request \emph{unsupported}.
If a model takes the modalities but fails, for example because the input exceeds its context, the request scores 0.

\subsection{Non-agentic decisions}
\label{sec:static}

This component contains \nAllQuestions{} questions across \nAllDomains{} domains (Table~\ref{tab:domains}):
\nImgQuestions{} image questions across \nImgTasks{} tasks, \nQuestions{} video questions across \nTasks{} tasks, and
\nAudQuestions{} audio questions across \nAudTasks{} tasks.
Each task covers one kind of decision and uses one or more public sources, except three tasks with
program-generated images or videos.
\nShortQuestions{} video questions use short videos, and \nLongQuestions{} use long videos; \nAVQuestions{} of them also
need the audio track, such as a meeting recording in which the model decides whether the group made a decision.
The domains cover basic perception, knowledge, professional decisions (sports fouls, driving risk, medical images,
machine sounds), retrieval, forecasting, media quality, model-as-judge, safety, embodied and robotics tasks, documents
and charts, GUI steps, games, dialogue, and routing.
Examples include deciding whether a meme is hateful, whether a clip shows a foul, whether a voice is spoofed, whether a
generated video matches its prompt, and whether to escalate a request to a slower model.
\nBenchTasks{} further video tasks hold 500-question subsets of public benchmarks with their original questions, and
\nImgStd{} further image tasks hold \nImgStdQuestions{} questions of standard image benchmarks
\citep{lu2022learn,Schwenk_2022,okvqa,gurari2018vizwiz,hudson2018gqa,yue2023mmmu,naturalbench}.
We report both separately.
Appendix~\ref{app:sources} lists all sources.

\begin{table}[t]
\centering
\footnotesize
\setlength{\tabcolsep}{3pt}
\begin{adjustbox}{max width=\columnwidth}
\begin{tabular}{@{}lrrrrrr@{}}
\toprule
\multirow{2}{*}{Domain} & \multicolumn{2}{c}{Image} & \multicolumn{2}{c}{Video}
& \multicolumn{2}{c}{Audio} \\
\cmidrule(lr){2-3} \cmidrule(lr){4-5} \cmidrule(lr){6-7}
& Tasks & Questions & Tasks & Questions & Tasks & Questions \\
\midrule
Basic perception & 3 & 2,000 & 5 & 2,500 & 6 & 5,079 \\
Knowledge and reasoning & 5 & 3,374 & 5 & 2,500 & 3 & 471 \\
Documents and charts & 3 & 1,597 & -- & -- & -- & -- \\
Social judgments & 2 & 1,000 & -- & -- & -- & -- \\
Professional & 4 & 2,850 & 3 & 1,500 & 2 & 1,301 \\
Retrieval & 2 & 1,000 & 2 & 1,000 & 1 & 500 \\
Forecasting & -- & -- & 2 & 1,000 & -- & -- \\
Batch processing & 1 & 1,000 & 1 & 500 & -- & -- \\
Model-as-judge & 4 & 1,886 & 3 & 1,500 & 2 & 1,409 \\
Safety & 5 & 2,295 & 1 & 500 & 2 & 667 \\
Embodied and robotics & 1 & 500 & 4 & 2,000 & -- & -- \\
GUI steps & 2 & 749 & -- & -- & -- & -- \\
Games & 3 & 1,749 & 1 & 500 & -- & -- \\
Code rendering & 1 & 500 & -- & -- & -- & -- \\
Dialogue & -- & -- & 1 & 500 & 3 & 1,133 \\
Routing & 4 & 2,500 & 2 & 1,000 & -- & -- \\
Public benchmark subsets & -- & -- & 5 & 2,500 & -- & -- \\
\midrule
All & 40 & 23,000 & 35 & 17,500 & 19 & 10,560 \\
\bottomrule
\end{tabular}
\end{adjustbox}
\caption{\textbf{Domains of the non-agentic decisions.} A task is one type of decision, and it draws on one or more
sources, most of them public. Video counts short videos, long videos, and audio-visual input together. The public
benchmark subsets keep their original questions and are reported apart from the other domains. --: no task.}
\label{tab:domains}
\end{table}

\paragraph{Labels and instructions.}
Every label comes from the original source, and we do not create new labels.
They include expert annotations, recorded outcomes, annotator scores, and other original dataset labels.
When a source gives instructions or annotation guidelines, we keep their wording or adapt it to the
decision-request format.
Otherwise, we write the instructions from the source's labeling criteria.
For the \nRecipeSourcesRelease{} sources whose annotations we do not redistribute, we release only how each question
is built, and a loader rebuilds the questions from the source's own annotations.

\subsection{Closed-loop games}
\label{sec:loop}

In a game, a model must choose a sequence of actions to reach a goal.
Each action changes the next screen, so success depends on the decisions of a whole episode.
We include \nEnvs{} environments from \nEnvSuites{} suites: browser games from GameWorld \cite{ouyang2026gameworld}, the
BabyAI, Baba Is AI, MiniHack, Crafter, NetHack, and TextWorld tasks of BALROG \cite{paglieri2025balrog}, and the
multi-agent games of VS-Bench \cite{xu2025vsbench} (Table~\ref{tab:loop}).
Each suite runs with its original harness.
A bridge converts each step into a decision request: the screenshots and the text observation form the state, and the
legal actions form one choice question.
The game executes the option with the highest probability.
All models use the same bridge and the same episodes.
We report the score of each suite: the task progress in GameWorld, the progression in BALROG, and the return in
VS-Bench, which we normalize per game between the lowest and the highest return of the models.
Each score averages the environments of its suite, each environment with equal weight.
\nPlayableEnvs{} of the \nEnvs{} environments are playable, while the others need free-text actions or do not run on the
pinned harness (Appendix~\ref{app:envs}).

\paragraph{Real-time mode.}
In paused mode, the game waits for each decision.
In real-time mode, the game advances by the model's measured median latency for each decision.
Slower models therefore act on older screens.

\begin{table}[t]
\centering
\footnotesize
\setlength{\tabcolsep}{4pt}
\begin{tabular}{@{}lrrrl@{}}
\toprule
Suite & Envs & Playable & Episodes & Score \\
\midrule
GameWorld & 34 & 33 & 170 & progress \\
BALROG & 58 & 54 & 255 & progression \\
VS-Bench & 10 & 10 & 100 & normalized return \\
\bottomrule
\end{tabular}
\caption{\textbf{Closed-loop game suites.} Playable: environments whose actions a choice question can express and
that run on the pinned harness. Episodes: per model, over all environments. All episodes are in paused mode.}
\label{tab:loop}
\end{table}

\subsection{Metrics}
\label{sec:metrics}

As in System1-Arena, we evaluate the probability distribution that a model returns for each question.
\textbf{Accuracy} compares the most probable option with the hard reference label.
The \textbf{Brier score} \cite{brier1950verification}, $\sum_o (p(o) - t(o))^2$, measures probability quality, with
$t$ a one-hot label or a distribution of annotator labels.
The score ranges from 0 to 2, and lower values indicate better probabilities.
A task score averages its sources, and a group score averages its tasks, each task with equal weight.
We average the image, video, and audio tasks separately.
The static score averages the \nStaticTasks{} video tasks that every model takes and leaves out the audio-visual tasks.
\textbf{Coverage} is the share of questions that a model can take; unsupported requests are left out of its scores.
\textbf{Latency} is the median time per request.

\section{Experimental Setup}
\label{sec:setup}

\paragraph{Probabilities from each model family.}
For each question $q$ with options $o_1, \dots, o_K$, every model provides a distribution $p(o_k \mid s, q)$.
\emph{System-1 models} return it directly, from a decision head or the logits of their option labels.

For \emph{open MLLMs}, we obtain probabilities from answer labels.
The prompt contains the state with its media, one question, and its options, which each carry a short label $\ell_k$.
Labels are A, B, \dots{} for choices, 0--9 for the levels of a score, and yes/no for boolean questions.
The prompt ends where the answer starts, and one forward pass gives $z_k = \log P_\theta(\ell_k \mid x)$ for each
label.
We normalize these values as $p(o_k \mid s, q) = \exp z_k / \sum_{j} \exp z_j$.
For DiffusionGemma \cite{diffusiongemma2026}, one read-only denoising step gives the label probabilities at the answer
slot.

\paragraph{Models.}
We evaluate \nModels{} open models: \nSystemOne{} System-1 models and \nOpenMLLMs{} open MLLMs.
For each System-1 model, we name its base MLLM, as its model card states.
\begin{itemize}[leftmargin=*,itemsep=1pt,topsep=2pt]
  \item \textbf{System-1 models}:
  d3 \cite{vllmsr2026d3}, Clef \cite{cloudflare2026clef}, and OneJev-27B \cite{omnijev2026onejev}, built on Qwen3.8-27B;
  Clef-Flash and OneJev-9B, built on Qwen3.5-9B; OneJev-4B and OneJev-0.8B, built on Qwen3.5 of the same size;
  Clef-Omni, built on Qwen3-Omni-30B-A3B-Instruct;
  Jev-Omni \cite{jevomni2026}, built on Gemma~4 12B, with its own decision head and a second head released separately
  \cite{ferdinandl2026jevomnihead};
  Xor-26B-A4B \cite{juspay2026xor}, built on Gemma~4 26B-A4B;
  and Raya-Decision-v1 \cite{raya2026decision}, built on Qwen3.5-2B.
  \item \textbf{Open MLLMs}:
  Qwen3.8-27B \cite{qwen2026qwen38} in FP8;
  Qwen3.5 \cite{qwen2026qwen35} with 0.8B, 2B, 4B, and 9B parameters, and 35B-A3B in GPTQ-Int4;
  Qwen3-VL-30B-A3B \cite{qwen2025qwen3vl};
  Qwen3-Omni-30B-A3B \cite{qwen2025qwen3omni}, Instruct and Thinking;
  Qwen2.5-Omni \cite{xu2025qwen25omni} with 3B and 7B parameters, the 7B also in AWQ and GPTQ;
  Gemma~4 12B and 26B-A4B \cite{google2026gemma4};
  DiffusionGemma-26B-A4B \cite{diffusiongemma2026}, also in AWQ;
  MiniCPM-o 2.6 and 4.5 \cite{openbmb2026minicpmo,cui2026minicpmo45};
  Phi-4-multimodal \cite{microsoft2025phi4mm};
  and Nemotron-3-Nano-Omni \cite{nvidia2026nemotronomni}.
\end{itemize}
All \nModels{} models take the image and video tasks, and the \nAudModels{} that take audio have audio results.
\textbf{Coverage} (Section~\ref{sec:metrics}) is 100\% for these and \nNoAudCoverage\% for the other
\nNoAudModels{}, which take neither audio nor the \nAVQuestions{} audio-visual questions.

\paragraph{Thinking.}
In the main results, every open MLLM runs with thinking off.
Qwen3-Omni-30B-A3B-Thinking and Nemotron-3-Nano-Omni think by default, so their prompts end with an empty thinking
block.
The \nThinkModels{} MLLMs that have a thinking mode also run with thinking on, with the thinking budgets of
System1-Arena \cite{chen2026system1arena}: each question gets one thought, which is sampled with the model's settings
and stopped at 2{,}048 tokens.
For a smaller budget, we cut the same thought, close it as the chat template does, and read the label probabilities as
above.
These runs are not counted among the \nModels{} models (Appendix~\ref{app:thinking}).

\paragraph{Reading policies.}
Each model reads the media as its release specifies.
OneJev reads every frame of a one-frame-per-second grid.
Jev-Omni reads 16 frames from that grid and the whole audio track.
Raya reads 8 frames; Xor, Gemma~4, and DiffusionGemma read 32 frames; and d3 reads 4 to 32 frames at two frames per
second.
Clef, Clef-Flash, and the Qwen models use their processors' defaults, from 4 to 768 frames at two frames per second.
Clef-Omni reads a one-frame-per-second grid fitted to its context.
MiniCPM-o and Nemotron-3-Nano-Omni read at most 64 frames at one frame per second, and Phi-4-multimodal reads 16 frames.
Jev-Omni, Clef-Omni, the Qwen omni models, Gemma~4 12B, MiniCPM-o, Phi-4-multimodal, and Nemotron-3-Nano-Omni also take
audio; some of them cut long audio tracks (Appendix~\ref{app:details}).

\paragraph{Protocol.}
Each request receives one sample.
Questions with failed calls, unusable responses, or rejected inputs receive no accuracy credit.
In games, a failed call ends the episode as a technical failure; the bridge does not substitute a default action.
Local models run on NVIDIA A100 GPUs (40~GB) and NVIDIA RTX A6000 GPUs (48~GB).

\section{Results}
\label{sec:results}

\begin{table*}[t]
\centering
\footnotesize
\setlength{\tabcolsep}{2.9pt}
\begin{adjustbox}{max width=\textwidth}
\begin{tabular}{@{}rl>{\columncolor{s1overall}}r>{\columncolor{s1overall}}rrrrrrrrrrr@{}}
\toprule
\multirow{2}{*}{Rank} & \multirow{2}{*}{Model} & \multicolumn{2}{c}{Static}
& \multicolumn{2}{c}{Short video} & \multicolumn{2}{c}{Long video}
& \multicolumn{2}{c}{Audio-visual} & Public & \multicolumn{3}{c}{Closed loop (paused)} \\
\cmidrule(lr){3-4} \cmidrule(lr){5-6} \cmidrule(lr){7-8} \cmidrule(lr){9-10} \cmidrule(lr){12-14}
& & Acc. & Brier & Acc. & Brier & Acc. & Brier & Acc. & Brier & subsets & GameWorld & BALROG & VS-Bench \\
\midrule
\rowcolor{s1band}\multicolumn{14}{@{}l}{\textit{System-1 models}} \\
\rowcolor{s1blueA}1 & d3 & \textbf{66.9} & 0.419 & \textbf{66.9} & \textbf{0.430} & 60.3 & 0.455 & -- & -- & 76.2 & \textbf{40.9} & 29.2 & \textbf{45.3} \\
\rowcolor{s1blueB}2 & Clef & 66.5 & 0.425 & 65.9 & 0.442 & 60.5 & 0.457 & -- & -- & \textbf{77.5} & 30.9 & \textbf{33.3} & 25.3 \\
4 & OneJev-27B & 65.8 & \textbf{0.417} & 64.4 & 0.438 & \textbf{61.7} & \textbf{0.428} & -- & -- & 76.8 & 39.8 & 22.0 & 32.4 \\
5 & OneJev-9B & 64.6 & 0.428 & 64.6 & 0.434 & 57.6 & 0.479 & -- & -- & 74.5 & 32.2 & 4.1 & 13.0 \\
7 & Clef-Flash & 63.3 & 0.464 & 63.3 & 0.476 & 55.3 & 0.514 & -- & -- & 74.7 & 30.1 & 4.8 & 20.8 \\
8 & OneJev-4B & 62.9 & 0.448 & 64.2 & 0.440 & 52.9 & 0.527 & -- & -- & 72.0 & 32.0 & 3.3 & 20.8 \\
12 & Clef-Omni & 60.2 & 0.508 & 60.1 & 0.528 & 53.2 & 0.536 & \textbf{69.5} & \textbf{0.455} & 70.3 & 24.6 & 1.8 & 23.9 \\
21 & Xor-26B-A4B & 54.7 & 0.534 & 55.3 & 0.530 & 47.4 & 0.579 & -- & -- & 62.5 & 25.1 & 6.4 & 30.3 \\
24 & OneJev-0.8B & 53.6 & 0.542 & 56.7 & 0.516 & 42.8 & 0.627 & -- & -- & 56.1 & 20.7 & 3.9 & 12.1 \\
27 & Jev-Omni (head v3) & 53.2 & 0.546 & 54.6 & 0.537 & 45.8 & 0.587 & 40.1 & 0.724 & 58.0 & 24.5 & 4.8 & 15.6 \\
29 & Jev-Omni & 52.9 & 0.575 & 54.8 & 0.556 & 44.8 & 0.635 & 45.5 & 0.685 & 56.2 & 25.8 & 5.6 & 19.7 \\
33 & Raya-Decision-v1 & 46.1 & 0.706 & 50.4 & 0.682 & 30.7 & 0.814 & -- & -- & 50.5 & 15.2 & 0.0 & 24.3 \\
\midrule
\rowcolor{s1band}\multicolumn{14}{@{}l}{\textit{Open MLLMs, read at the answer token}} \\
\rowcolor{s1blueC}3 & Qwen3.5-35B-A3B (Int4) & 66.0 & 0.427 & 65.5 & 0.442 & 59.4 & 0.461 & -- & -- & 77.3 & 31.2 & 7.0 & 25.0 \\
6 & Qwen3.5-9B & 63.9 & 0.457 & 63.9 & 0.463 & 55.7 & 0.521 & -- & -- & 75.2 & 29.5 & 3.7 & 24.8 \\
9 & Qwen3-VL-30B-A3B & 62.9 & 0.512 & 62.7 & 0.521 & 55.4 & 0.568 & -- & -- & 74.3 & 23.4 & 2.3 & 24.0 \\
10 & Qwen3.8-27B (FP8) & 62.5 & 0.509 & 62.9 & 0.518 & 53.9 & 0.590 & -- & -- & 73.2 & 34.6 & 8.1 & 19.9 \\
11 & Qwen3.5-4B & 62.0 & 0.478 & 62.8 & 0.480 & 53.4 & 0.537 & -- & -- & 70.5 & 31.8 & 3.4 & 27.6 \\
13 & MiniCPM-o 4.5 & 60.0 & 0.543 & 60.8 & 0.544 & 53.8 & 0.582 & 55.2 & 0.742 & 65.6 & 21.6 & 2.4 & 24.6 \\
14 & Qwen3-Omni-30B-A3B (Thinking) & 59.6 & 0.617 & 62.0 & 0.582 & 46.8 & 0.781 & 59.8 & 0.628 & 67.7 & 20.5 & 2.5 & 17.0 \\
15 & Nemotron-3-Nano-Omni & 58.7 & 0.546 & 59.7 & 0.542 & 48.9 & 0.626 & 65.5 & 0.531 & 68.0 & 18.8 & 5.3 & 22.0 \\
16 & Qwen3-Omni-30B-A3B & 58.2 & 0.532 & 58.5 & 0.526 & 50.8 & 0.609 & 55.3 & 0.562 & 67.7 & 24.6 & 2.3 & 35.2 \\
17 & Gemma~4 26B-A4B & 56.8 & 0.777 & 57.9 & 0.771 & 50.0 & 0.864 & -- & -- & 61.8 & 32.5 & 9.7 & 22.2 \\
18 & Qwen2.5-Omni-7B & 55.5 & 0.555 & 56.7 & 0.548 & 44.9 & 0.638 & 57.6 & 0.653 & 65.4 & 23.8 & 3.3 & 25.1 \\
19 & Qwen2.5-Omni-7B (AWQ) & 55.4 & 0.550 & 56.9 & 0.540 & 44.6 & 0.631 & 57.5 & 0.637 & 64.6 & 21.2 & 2.6 & 28.6 \\
20 & Qwen2.5-Omni-7B (GPTQ) & 55.2 & 0.551 & 57.3 & 0.529 & 42.5 & 0.661 & 56.4 & 0.632 & 64.6 & 23.4 & 2.4 & 28.8 \\
22 & DiffusionGemma-26B-A4B (AWQ) & 54.2 & 0.663 & 56.1 & 0.646 & 45.5 & 0.758 & -- & -- & 58.9 & 25.6 & 4.1 & 9.2 \\
23 & DiffusionGemma-26B-A4B & 54.1 & 0.651 & 55.8 & 0.643 & 45.8 & 0.732 & -- & -- & 59.2 & 25.9 & 6.2 & 11.7 \\
25 & Qwen2.5-Omni-3B & 53.5 & 0.558 & 55.1 & 0.546 & 43.1 & 0.644 & 55.1 & 0.608 & 61.3 & 17.2 & 4.3 & 21.5 \\
26 & MiniCPM-o 2.6 & 53.4 & 0.669 & 55.3 & 0.656 & 43.1 & 0.758 & 52.1 & 0.639 & 60.5 & 18.3 & 0.5 & 17.5 \\
28 & Qwen3.5-2B & 53.0 & 0.555 & 54.8 & 0.542 & 41.7 & 0.635 & -- & -- & 62.0 & 13.6 & 1.3 & 15.9 \\
30 & Gemma~4 12B & 51.8 & 0.790 & 53.1 & 0.794 & 45.4 & 0.833 & 46.4 & 0.964 & 55.3 & 25.1 & 7.5 & 40.4 \\
31 & Phi-4-multimodal & 49.7 & 0.655 & 52.6 & 0.627 & 38.7 & 0.738 & 58.8 & 0.542 & 53.3 & 18.0 & 2.2 & 19.8 \\
32 & Qwen3.5-0.8B & 47.9 & 0.632 & 49.3 & 0.619 & 38.8 & 0.709 & -- & -- & 54.9 & 18.5 & 0.0 & 9.6 \\
\bottomrule
\end{tabular}
\end{adjustbox}
\caption{\textbf{Video and closed-loop results of 33 models.} Accuracy (\%) and Brier score average
the tasks of each input kind, each task with equal weight: \nShortTasks{} short-video, \nLongTasks{} long-video and
\nAVTasks{} audio-visual tasks, and the \nBenchTasks{} public benchmark subsets (accuracy only). Static averages
the \nStaticTasks{} video tasks that every model takes (short video, long video, and public subsets) and leaves out
the audio-visual tasks. Rank is by static accuracy, and within a family the rows follow it. The grey columns are the static
scores, and a lower Brier score is better. Blue rows are the three most accurate models (darker is better), and bold is
the best per column. --: the model does not take audio. Closed loop: GameWorld progress and BALROG progression (\%), mean
over \nPlayableEnvs{} playable environments; VS-Bench return, min-max normalized per game over the models with results
(\%).}
\label{tab:main}
\end{table*}

Tables~\ref{tab:main} and~\ref{tab:imgaud} report accuracy and Brier score per input kind and the closed-loop scores
of \nModels{} open models.
Table~\ref{tab:pairs} compares each System-1 model with its base MLLM.
Non-agentic tasks pose decisions on recorded inputs, and agentic tasks pose them in closed loops and conversations.

\paragraph{Non-agentic: System-1 models lead on images.}
The three most accurate models on the image tasks are System-1 models, led by d3 at 69.7\%, compared with 64.8\% for
the best MLLM, Qwen3.8-27B (Table~\ref{tab:imgaud}).
On static video and audio, the best System-1 model leads the best MLLM by only 0.9 and 0.4 points, and the four most
accurate models on static video are within 1.1 points of each other.
System-1 models lead more on long videos and on audio-visual input, by 2.3 and 4.0 points.
These MLLMs answer in one forward pass at the answer token, without reasoning.
Per task, the best System-1 model leads the best MLLM most on escalating a request to a slower model, on the needle
task, and on the direction of motion (Appendix Tables~\ref{tab:leaves} and~\ref{tab:leaves_mllm}).

\begin{table}[t]
\centering
\scriptsize
\setlength{\tabcolsep}{2.0pt}
\begin{tabular}{@{}lrrrrrrr@{}}
\toprule
\multirow{2}{*}{Model} & \multicolumn{4}{c}{Image} & \multicolumn{3}{c}{Audio} \\
\cmidrule(lr){2-5} \cmidrule(lr){6-8}
& Acc. & Brier & ECE & Public & Acc. & Brier & ECE \\
\midrule
\rowcolor{s1band}\multicolumn{8}{@{}l}{\textit{System-1 models}} \\
\rowcolor{s1blueA}d3 & \textbf{69.7} & \textbf{0.377} & 0.095 & \textbf{87.2} & -- & -- & -- \\
\rowcolor{s1blueC}Clef & 67.3 & 0.416 & 0.115 & 85.1 & -- & -- & -- \\
\rowcolor{s1blueB}OneJev-27B & 67.8 & 0.388 & \textbf{0.077} & 85.9 & -- & -- & -- \\
OneJev-9B & 64.5 & 0.426 & 0.084 & 84.0 & -- & -- & -- \\
Clef-Flash & 62.8 & 0.465 & 0.113 & 83.7 & -- & -- & -- \\
OneJev-4B & 62.4 & 0.444 & 0.082 & 83.6 & -- & -- & -- \\
\rowcolor{s1greenA}Clef-Omni & 62.0 & 0.488 & 0.139 & 83.9 & \textbf{62.9} & 0.495 & 0.241 \\
Xor-26B-A4B & 64.7 & 0.439 & 0.134 & 83.8 & -- & -- & -- \\
OneJev-0.8B & 51.8 & 0.548 & 0.101 & 72.3 & -- & -- & -- \\
Jev-Omni (head v3) & 55.1 & 0.543 & 0.148 & 78.1 & 48.4 & 0.541 & \textbf{0.132} \\
Jev-Omni & 57.0 & 0.546 & 0.166 & 77.3 & 47.7 & 0.568 & 0.179 \\
Raya-Decision-v1 & 41.2 & 0.739 & 0.232 & 65.8 & -- & -- & -- \\
\midrule
\rowcolor{s1band}\multicolumn{8}{@{}l}{\textit{Open MLLMs, read at the answer token}} \\
Qwen3.5-35B-A3B (Int4) & 64.5 & 0.461 & 0.148 & 84.8 & -- & -- & -- \\
Qwen3.5-9B & 60.7 & 0.492 & 0.150 & 82.0 & -- & -- & -- \\
Qwen3-VL-30B-A3B & 58.8 & 0.599 & 0.236 & 83.2 & -- & -- & -- \\
Qwen3.8-27B (FP8) & 64.8 & 0.439 & 0.112 & 83.5 & -- & -- & -- \\
Qwen3.5-4B & 57.3 & 0.521 & 0.136 & 81.0 & -- & -- & -- \\
MiniCPM-o 4.5 & 59.3 & 0.563 & 0.207 & 81.4 & 57.9 & 0.548 & 0.237 \\
\rowcolor{s1greenC}Qwen3-Omni-30B-A3B (Thinking) & 57.5 & 0.666 & 0.289 & 82.1 & 59.8 & 0.570 & 0.274 \\
Nemotron-3-Nano-Omni & 54.6 & 0.629 & 0.236 & 83.3 & 54.4 & 0.573 & 0.231 \\
\rowcolor{s1greenB}Qwen3-Omni-30B-A3B & 54.3 & 0.672 & 0.278 & 81.0 & 62.5 & 0.520 & 0.235 \\
Gemma~4 26B-A4B & 64.5 & 0.629 & 0.318 & 82.6 & -- & -- & -- \\
Qwen2.5-Omni-7B & 55.4 & 0.568 & 0.170 & 79.8 & 57.0 & 0.479 & 0.154 \\
Qwen2.5-Omni-7B (AWQ) & 54.5 & 0.570 & 0.168 & 78.9 & 56.7 & 0.480 & 0.151 \\
Qwen2.5-Omni-7B (GPTQ) & 54.7 & 0.560 & 0.153 & 79.1 & 55.5 & 0.499 & 0.163 \\
DiffusionGemma-26B-A4B (AWQ) & 60.7 & 0.579 & 0.244 & 81.8 & -- & -- & -- \\
DiffusionGemma-26B-A4B & 61.9 & 0.543 & 0.217 & 82.1 & -- & -- & -- \\
Qwen2.5-Omni-3B & 49.6 & 0.583 & 0.135 & 76.4 & 58.1 & \textbf{0.476} & 0.145 \\
MiniCPM-o 2.6 & 53.8 & 0.615 & 0.220 & 79.5 & 54.4 & 0.583 & 0.244 \\
Qwen3.5-2B & 46.6 & 0.647 & 0.199 & 75.6 & -- & -- & -- \\
Gemma~4 12B & 61.6 & 0.647 & 0.316 & 78.9 & 53.0 & 0.730 & 0.359 \\
Phi-4-multimodal & 37.0 & 0.872 & 0.384 & 55.8 & 43.7 & 0.716 & 0.273 \\
Qwen3.5-0.8B & 38.8 & 0.695 & 0.210 & 64.5 & -- & -- & -- \\
\bottomrule
\end{tabular}
\caption{\textbf{Image and audio results of 33 models.} Image: \nImgTasks{} tasks from \nImgSources{} sources
(\nImgQuestions{} questions); audio: \nAudTasks{} tasks from \nAudSources{} sources (\nAudQuestions{}
questions). Each score averages the tasks, each task with equal weight. Public: the mean accuracy on \nImgStd{} standard image benchmarks (\nImgStdQuestions{}
questions), reported apart like the public subsets of the video tasks. A lower Brier score and expected calibration
error (ECE, ten equal-mass bins) are better. Blue rows are the three most accurate models on the image tasks, and green
rows the three most accurate on the audio tasks (darker is better), and bold is the best per column. --: the model
does not take audio.}
\label{tab:imgaud}
\end{table}

\begin{table}[t]
\centering
\scriptsize
\setlength{\tabcolsep}{2.2pt}
\begin{adjustbox}{max width=\columnwidth}
\begin{tabular}{@{}lrrrrrrrrrrr@{}}
\toprule
\multirow{3}{*}{System-1 model}
& \multicolumn{8}{c}{Non-agentic} & \multicolumn{3}{c}{Agentic} \\
\cmidrule(lr){2-9} \cmidrule(lr){10-12}
& \multicolumn{2}{c}{Video static} & \multicolumn{2}{c}{AV} & \multicolumn{2}{c}{Image}
& \multicolumn{2}{c}{Audio} & \multicolumn{3}{c}{Closed loop} \\
\cmidrule(lr){2-3} \cmidrule(lr){4-5} \cmidrule(lr){6-7} \cmidrule(lr){8-9} \cmidrule(lr){10-12}
& Acc. & Brier & Acc. & Brier & Acc. & Brier & Acc. & Brier & GameWorld & BALROG & VS-Bench \\
\midrule
d3 & \textcolor{s1gain}{+4.4} & \textcolor{s1gain}{$-$0.090} & -- & -- & \textcolor{s1gain}{+4.9} & \textcolor{s1gain}{$-$0.062} & -- & -- & \textcolor{s1gain}{+6.4} & \textcolor{s1gain}{+21.1} & \textcolor{s1gain}{+25.4} \\
\hspace{0.5em}\textcolor{gray}{from Qwen3.8-27B (FP8)} & & & & & & & & & & & \\
Clef & \textcolor{s1gain}{+4.0} & \textcolor{s1gain}{$-$0.085} & -- & -- & \textcolor{s1gain}{+2.5} & \textcolor{s1gain}{$-$0.023} & -- & -- & \textcolor{s1loss}{$-$3.7} & \textcolor{s1gain}{+25.2} & \textcolor{s1gain}{+5.4} \\
\hspace{0.5em}\textcolor{gray}{from Qwen3.8-27B (FP8)} & & & & & & & & & & & \\
OneJev-27B & \textcolor{s1gain}{+3.2} & \textcolor{s1gain}{$-$0.092} & -- & -- & \textcolor{s1gain}{+3.0} & \textcolor{s1gain}{$-$0.051} & -- & -- & \textcolor{s1gain}{+5.2} & \textcolor{s1gain}{+13.9} & \textcolor{s1gain}{+12.4} \\
\hspace{0.5em}\textcolor{gray}{from Qwen3.8-27B (FP8)} & & & & & & & & & & & \\
OneJev-9B & +0.7 & \textcolor{s1gain}{$-$0.029} & -- & -- & \textcolor{s1gain}{+3.8} & \textcolor{s1gain}{$-$0.066} & -- & -- & \textcolor{s1gain}{+2.7} & +0.5 & \textcolor{s1loss}{$-$11.9} \\
\hspace{0.5em}\textcolor{gray}{from Qwen3.5-9B} & & & & & & & & & & & \\
Clef-Flash & $-$0.6 & +0.007 & -- & -- & \textcolor{s1gain}{+2.1} & \textcolor{s1gain}{$-$0.027} & -- & -- & +0.6 & \textcolor{s1gain}{+1.1} & \textcolor{s1loss}{$-$4.0} \\
\hspace{0.5em}\textcolor{gray}{from Qwen3.5-9B} & & & & & & & & & & & \\
OneJev-4B & +0.9 & \textcolor{s1gain}{$-$0.030} & -- & -- & \textcolor{s1gain}{+5.1} & \textcolor{s1gain}{$-$0.077} & -- & -- & +0.2 & $-$0.1 & \textcolor{s1loss}{$-$6.8} \\
\hspace{0.5em}\textcolor{gray}{from Qwen3.5-4B} & & & & & & & & & & & \\
Clef-Omni & \textcolor{s1gain}{+1.9} & \textcolor{s1gain}{$-$0.024} & \textcolor{s1gain}{+14.2} & \textcolor{s1gain}{$-$0.107} & \textcolor{s1gain}{+7.7} & \textcolor{s1gain}{$-$0.184} & +0.4 & \textcolor{s1gain}{$-$0.025} & +0.0 & $-$0.4 & \textcolor{s1loss}{$-$11.2} \\
\hspace{0.5em}\textcolor{gray}{from Qwen3-Omni-30B-A3B} & & & & & & & & & & & \\
Xor-26B-A4B & \textcolor{s1loss}{$-$2.1} & \textcolor{s1gain}{$-$0.243} & -- & -- & +0.2 & \textcolor{s1gain}{$-$0.190} & -- & -- & \textcolor{s1loss}{$-$7.4} & \textcolor{s1loss}{$-$3.3} & \textcolor{s1gain}{+8.1} \\
\hspace{0.5em}\textcolor{gray}{from Gemma~4 26B-A4B} & & & & & & & & & & & \\
OneJev-0.8B & \textcolor{s1gain}{+5.6} & \textcolor{s1gain}{$-$0.090} & -- & -- & \textcolor{s1gain}{+13.0} & \textcolor{s1gain}{$-$0.147} & -- & -- & \textcolor{s1gain}{+2.2} & \textcolor{s1gain}{+3.9} & \textcolor{s1gain}{+2.5} \\
\hspace{0.5em}\textcolor{gray}{from Qwen3.5-0.8B} & & & & & & & & & & & \\
Jev-Omni (head v3) & \textcolor{s1gain}{+1.4} & \textcolor{s1gain}{$-$0.244} & \textcolor{s1loss}{$-$6.3} & \textcolor{s1gain}{$-$0.240} & \textcolor{s1loss}{$-$6.5} & \textcolor{s1gain}{$-$0.104} & \textcolor{s1loss}{$-$4.6} & \textcolor{s1gain}{$-$0.189} & $-$0.6 & \textcolor{s1loss}{$-$2.7} & \textcolor{s1loss}{$-$24.8} \\
\hspace{0.5em}\textcolor{gray}{from Gemma~4 12B} & & & & & & & & & & & \\
Jev-Omni & \textcolor{s1gain}{+1.1} & \textcolor{s1gain}{$-$0.216} & $-$0.9 & \textcolor{s1gain}{$-$0.278} & \textcolor{s1loss}{$-$4.6} & \textcolor{s1gain}{$-$0.101} & \textcolor{s1loss}{$-$5.3} & \textcolor{s1gain}{$-$0.162} & +0.7 & \textcolor{s1loss}{$-$1.9} & \textcolor{s1loss}{$-$20.6} \\
\hspace{0.5em}\textcolor{gray}{from Gemma~4 12B} & & & & & & & & & & & \\
Raya-Decision-v1 & \textcolor{s1loss}{$-$6.9} & \textcolor{s1loss}{+0.152} & -- & -- & \textcolor{s1loss}{$-$5.4} & \textcolor{s1loss}{+0.092} & -- & -- & \textcolor{s1gain}{+1.5} & \textcolor{s1loss}{$-$1.3} & \textcolor{s1gain}{+8.4} \\
\hspace{0.5em}\textcolor{gray}{from Qwen3.5-2B} & & & & & & & & & & & \\
\bottomrule
\end{tabular}
\end{adjustbox}
\caption{\textbf{System-1 models minus the MLLMs they are post-trained from.} Each System-1 model is followed by
its base MLLM in gray. Differences in percentage points (Brier: absolute). Positive differences in accuracy and in
closed-loop scores, and negative differences in Brier score, favor the System-1 model. Video static: the
\nStaticTasks{} video tasks that every model takes. AV: audio-visual tasks. Image and audio: the tasks of
Table~\ref{tab:imgaud}. Agentic: the paused closed-loop scores of Table~\ref{tab:main}. Green: the System-1 model
is better by at least 1 point (0.01 in Brier score), orange: worse by at least as much. --: one of the two models has
no result.}
\label{tab:pairs}
\end{table}

\paragraph{Non-agentic: System-1 models give better probabilities.}
Post-training for decisions mainly improves the probabilities (Table~\ref{tab:pairs}).
It lowers the Brier score in 10 of the 12 pairs on static video, in 11 on images, and in all three pairs on audio,
while the accuracy moves in both directions.
The largest reductions on static video come from the two Jev-Omni models and Xor-26B-A4B, whose accuracy changes by
at most 2.1 points.
The gap holds at equal accuracy (Figure~\ref{fig:front}b): OneJev-4B and Qwen3.8-27B are about equally accurate on
static video, but their Brier scores are 0.448 and 0.509.
OneJev-27B and d3 have the lowest Brier scores on static video, and the four OneJev models and d3 have the lowest
calibration errors on images (Table~\ref{tab:imgaud}).

\begin{figure}[t]
\centering
\includegraphics[width=\columnwidth]{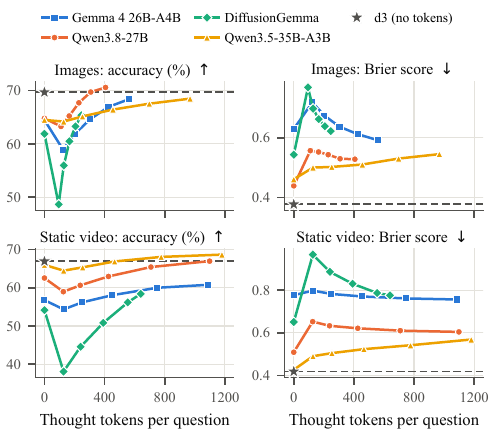}
\caption{\textbf{Thinking can raise accuracy but worsen Brier scores.} Accuracy and Brier score of four MLLMs on the
image tasks and the \nStaticTasks{} static video tasks against the mean number of thought tokens per question. At
zero tokens, thinking is off, as in Tables~\ref{tab:main} and~\ref{tab:imgaud}. The other points cut one thought
per question at 128 to 2{,}048 tokens. The star and the dashed line mark d3, which writes no tokens.}
\label{fig:thinking}
\end{figure}

\paragraph{Non-agentic: thinking improves accuracy, but not the probabilities.}
Figure~\ref{fig:thinking} gives four MLLMs thought budgets of at most 2{,}048 tokens.
With thinking off, all four are less accurate than d3, and a small budget can lower their accuracy: at 128 tokens,
DiffusionGemma loses more than 13 points on images and on video.
Larger budgets help, and at 2{,}048 tokens Qwen3.8-27B passes d3 on images and Qwen3.5-35B-A3B on video, with about
400 and 1{,}200 thought tokens per question.
Qwen3-Omni-30B-A3B (Thinking) also passes the best System-1 model on audio-visual and audio input.
The probabilities do not follow: thinking raises the Brier score of 10 of the \nThinkModels{} MLLMs on images and of 12
on video, and every Brier score with thinking stays above d3's (Appendix Table~\ref{tab:thinking}).

\paragraph{Non-agentic: brief evidence needs dense reading.}
OneJev-27B reads one frame per second, while Jev-Omni reads 16 frames regardless of the video length.
On long videos, OneJev-27B is 17.0 points more accurate.
The gap is largest when the evidence is brief: on the needle task, which asks whether and where a short clip was
inserted, every model above 80\% reads up to hundreds of frames.
With only 16 to 64 frames, OneJev-27B itself falls from 94.2\% to 60.2--84.6\% on this task, while its other
long-video tasks change little.
Dense reading costs latency, 2.17~s per long-video request for OneJev-27B against 0.90~s for Jev-Omni
(Table~\ref{tab:latency}).
A fixed frame budget therefore suits decisions about the whole clip, not those about a short moment.

\paragraph{Non-agentic: physics and counting remain hard.}
No model exceeds 61\% on balanced yes/no questions about physical outcomes, such as object contact in Physion and
collision risk in dashcam videos, where chance is 50\%.
Averaged over the models, both tasks stay near 52\%.
Repetition counting, which asks how many times an action repeats, is also hard: no model chooses the right count range
for more than 28.6\% of the questions.
In contrast, the models match a video to its caption with 96.2\% accuracy on average
(Appendix Tables~\ref{tab:leaves} and~\ref{tab:leaves_mllm}).
These decisions depend on dynamics over time rather than on the appearance of single frames.
Thinking does not close these gaps: at any budget, no MLLM exceeds 58\% on either physics task, and the best counting
accuracy rises only to 30.2\%.

\paragraph{Agentic: closed loops separate the models.}
Closed loops separate the models more than static decisions do (Figure~\ref{fig:front}a).
In BALROG, the three System-1 models built on Qwen3.8-27B reach 22.0--33.3\% progression, while their base reaches
8.1\% and no other model exceeds 9.7\% with thinking off.
Against their bases (Table~\ref{tab:pairs}), only these three gain substantially in BALROG, by 13.9 to 25.2 points,
while the other nine System-1 models change by $-3.3$ to $+3.9$ points.
In GameWorld, d3 and OneJev-27B are the two best models and lead their base by more than 5 points, while most other
pairs differ by less than 4 points.
In VS-Bench, d3 leads, but six of the other nine System-1 models score below their bases, the two Jev-Omni models by
more than 20 points.
Errors may compound in a loop: a static decision starts from a recorded state, whereas each decision in a loop changes
the states that follow.
With thinking on, Qwen3.8-27B itself reaches 28.0\% in BALROG and 49.3\% in GameWorld, but it writes 1{,}636 thought
tokens per decision on average (Appendix Table~\ref{tab:thinking}).
All episodes are paused, so latency does not change these scores.

\paragraph{Agentic: latency costs little in real time.}
In the real-time mode of GameWorld, each decision costs the model's median latency from a separate run on H100 PCIe
GPUs, 0.09 to 1.14~s.
Across models, progress changes by $-6.7$ to $+6.1$ points against paused play, and only Qwen3-Omni-30B-A3B (Thinking)
changes by more than two standard errors, with a gain (Appendix Table~\ref{tab:realtime}).
The change is only weakly related to latency, and the two slowest models, Clef and OneJev-27B, lose less than 3 points.
Slower hardware or single slow decisions could still cost progress that these runs do not show
(Appendix~\ref{app:limitations}).

\paragraph{Agentic: offline accuracy does not carry over to conversation.}
In four streaming environments built from Full-Duplex-Bench \cite{lin2025fdb_v1,lin2025fdb_v15}, a model decides at
each pause in the user's speech whether to speak, wait, continue, or stop, and one shared speech model speaks for it.
Offline, Clef-Omni tells a backchannel from an interruption in 99.7\% of the questions on the same recordings, and its
base Qwen3-Omni-30B-A3B in 74.8\%.
In the loop, the order reverses: Clef-Omni stops speaking in 77\% of the episodes in which the user only backchannels,
compared with 17\% for its base.
Both Jev-Omni models also stop more often than their base, Gemma~4 12B (Appendix~\ref{app:stream}).

\paragraph{Latency and cost.}
\begin{table}[t]
\centering
\footnotesize
\setlength{\tabcolsep}{2.5pt}
\begin{adjustbox}{max width=\columnwidth}
\begin{tabular}{@{}lrrrrrr@{}}
\toprule
\multirow{2}{*}{Model} & \multirow{2}{*}{GPUs}
& \multicolumn{5}{c}{Median latency per request (s)} \\
\cmidrule(lr){3-7}
& & Short & Long & AV & Public & Loop \\
\midrule
\rowcolor{s1band}\multicolumn{7}{@{}l}{\textit{System-1 models}} \\
d3 & 1 & 0.51 & 0.73 & -- & 1.14 & 0.54 \\
Clef & 1 & 1.28 & 3.15 & -- & 3.41 & 0.78 \\
OneJev-27B & 1 & 0.79 & 2.17 & -- & 2.26 & 0.76 \\
OneJev-9B & 1 & 0.47 & 1.19 & -- & 1.38 & 0.34 \\
Clef-Flash & 1 & 0.69 & 1.65 & -- & 1.83 & 0.35 \\
OneJev-4B & 1 & 0.43 & 1.10 & -- & 1.23 & 0.28 \\
Clef-Omni & 1 & 0.35 & 1.23 & 20.4 & 1.10 & 0.30 \\
Xor-26B-A4B & 1 & 0.79 & 0.88 & -- & 1.34 & 0.43 \\
OneJev-0.8B & 1 & 0.37 & 0.86 & -- & 1.00 & 0.19 \\
Jev-Omni (head v3) & 1 & 0.55 & 0.90 & 14.4 & 1.00 & 0.15 \\
Jev-Omni & 1 & 0.54 & 0.90 & 14.4 & 0.99 & 0.15 \\
Raya-Decision-v1 & 1 & \textbf{0.31} & \textbf{0.23} & -- & \textbf{0.35} & 0.22 \\
\midrule
\rowcolor{s1band}\multicolumn{7}{@{}l}{\textit{Open MLLMs, read at the answer token}} \\
Qwen3.5-35B-A3B (Int4) & 1 & 0.48 & 1.00 & -- & 1.19 & 0.20 \\
Qwen3.5-9B & 1 & 0.46 & 1.05 & -- & 1.24 & 0.18 \\
Qwen3-VL-30B-A3B & 1 & 0.43 & 1.03 & -- & 1.23 & 0.17 \\
Qwen3.8-27B (FP8) & 1 & 0.68 & 1.42 & -- & 1.64 & 0.30 \\
Qwen3.5-4B & 1 & 0.39 & 0.91 & -- & 1.09 & 0.14 \\
MiniCPM-o 4.5 & 1 & 0.43 & 1.33 & 4.04 & 1.29 & 0.38 \\
Qwen3-Omni-30B-A3B (Thinking) & 2 & 0.55 & 1.77 & 7.69 & 1.53 & 0.19 \\
Nemotron-3-Nano-Omni & 1 & 0.40 & 0.90 & \textbf{3.03} & 0.92 & 0.19 \\
Qwen3-Omni-30B-A3B & 2 & 0.55 & 1.77 & 7.56 & 1.53 & 0.19 \\
Gemma~4 26B-A4B & 1 & 0.65 & 0.70 & -- & 0.94 & 0.19 \\
Qwen2.5-Omni-7B & 1 & 0.61 & 1.99 & 5.40 & 1.83 & 0.26 \\
Qwen2.5-Omni-7B (AWQ) & 1 & 0.62 & 2.02 & 5.47 & 1.84 & 0.27 \\
Qwen2.5-Omni-7B (GPTQ) & 1 & 0.62 & 1.97 & 5.41 & 1.81 & 0.26 \\
DiffusionGemma-26B-A4B (AWQ) & 1 & 0.65 & 0.72 & -- & 0.92 & 0.20 \\
DiffusionGemma-26B-A4B & 1 & 0.64 & 0.72 & -- & 0.93 & 0.20 \\
Qwen2.5-Omni-3B & 1 & 0.56 & 1.78 & 5.07 & 1.71 & 0.22 \\
MiniCPM-o 2.6 & 1 & 0.43 & 1.29 & 3.49 & 1.27 & 0.37 \\
Qwen3.5-2B & 1 & 0.36 & 0.81 & -- & 0.98 & 0.12 \\
Gemma~4 12B & 1 & 0.61 & 0.65 & 3.17 & 0.80 & 0.13 \\
Phi-4-multimodal & 1 & 1.44 & 1.01 & 3.59 & 1.53 & 0.20 \\
Qwen3.5-0.8B & 1 & 0.33 & 0.74 & -- & 0.90 & \textbf{0.11} \\
\bottomrule
\end{tabular}
\end{adjustbox}
\caption{\textbf{Latency per decision at batch size 1.} Median wall time from a request to its probabilities on
NVIDIA H100 80\,GB SXM GPUs, one request at a time after a warm-up, including media decoding and every question of the
request; loading is not included. Every model gets the same requests: 10 per short-video, long-video and public-subset
task, 20 per audio-visual (AV) task, and 100 closed-loop decisions (Loop). --: the model does not take audio. Bold: the
lowest latency of a column.}
\label{tab:latency}
\end{table}

No model writes any output tokens before it answers (Table~\ref{tab:latency}): the System-1 models read their answer
from a decision head or the label logits, and the open MLLMs are read at the answer token with thinking off.
A decision therefore costs media decoding and the forward pass over the prompt, but no decoding of text.
On one H100, the median closed-loop decision takes from 0.11~s (Qwen3.5-0.8B) to 0.78~s (Clef).
OneJev-27B takes 0.79~s per short-video request, and Jev-Omni 0.54~s.
The latency runs use the same models and inputs as the accuracy runs and change only how a model is placed on the GPUs.

\section{Conclusion}
\label{sec:conclusion}

\name{} asks one question of every multimodal decision: can we leave it to a fast model that reads the media and
answers with probabilities in a single forward pass?
On static decisions, the answers are encouraging: System-1 models lead the MLLMs on images, are close on video and
audio, give better probabilities than their base MLLMs, and stay within 1.8 points of MLLMs that think for hundreds of
tokens.
They are sobering where a decision unfolds over time: post-training helps closed-loop play in only three of twelve
pairs, System-1 models stop for most backchannels in conversation, and no model predicts physical outcomes well.
Speed and reading are part of the answer: in real-time GameWorld, latencies of up to 1.1~s per decision cost no
measurable progress, while a fixed frame budget misses brief evidence in long videos.
We hope that the benchmark helps decide, for each decision, whether reading more of the media or thinking longer is
worth its price.

{
    \small
    \bibliographystyle{ieeenat_fullname}
    \bibliography{references,datasets}
}

\clearpage
\twocolumn[\centering\Large\textbf{\thetitle}\\\vspace{0.5em}Appendix\\\vspace{1.0em}]
\appendix

\section{Prompts}
\label{app:prompt}
Figure~\ref{fig:prompt} shows the prompt of the \emph{open MLLMs} for one request (Section~\ref{sec:setup}), which
follows the prompt of System1-Arena~\cite{chen2026system1arena}.
Each medium is sent before the text and replaced in the state by a placeholder such as \texttt{<video 1>}, and questions
with more than 26 options are labeled with zero-padded numbers.
The \emph{System-1 models} take no prompt of ours.
Each receives the request in the format of its model card: the state with each medium replaced by its placeholder, and
the questions with their types and options.
Its own code turns this request into the model input, as its card runs it: the official engine for OneJev, the
repository's helper for Jev-Omni, the joint-schema encoder for Clef, the runtime of d3, and the served wrapper for Xor.
Raya has no such interface, and we write the prompt that its card gives: the media, the text, the question, the options
as labeled lines, and an instruction to answer with one label.

\begin{figure}[t]
\centering
\fbox{\parbox{0.95\columnwidth}{\scriptsize
\textit{System:} \texttt{You are a classifier. Read the input and answer the question. Reply with the label of your
answer only.}\\[4pt]
\textit{User}, after the video:\\
\ttfamily
Media, in the order shown: "<video 1>"\\[4pt]
Input:\\
\{\\
\ "text": "Observe the provided video.",\\
\ "video": "<video 1>"\\
\}\\[4pt]
Question: Which action category best describes the main action in this video?\\
Options:\\
A. eating hotdog\\
B. eating nachos\\
C. eating ice cream\\
D. yawning\\
Reply with the label of one option.}}
\caption{\textbf{The prompt of an open MLLM for one request} (Kinetics-700, action candidates).}
\label{fig:prompt}
\end{figure}

\section{Evaluation Details}
\label{app:details}

\paragraph{Unsupported and rejected requests.}
Before any call, the runner compares the media kinds of a request with the modalities of the model.
If one is missing, it stores the request as unsupported and makes no call.
Unsupported requests are left out of the model's scores.
A request that a model takes but cannot answer, for example because the input exceeds its context, scores 0.

\paragraph{Inputs cut by memory limits.}
On 40~GB GPUs, some inputs are cut, as the release of each model allows.
Nemotron-3-Nano-Omni hears the first 600~s of audio and sees at most 64 frames.
Phi-4-multimodal hears the first 640~s, MiniCPM-o 4.5 and 2.6 the first 800~s and 320~s, and Qwen2.5-Omni the first
300~s, as its processor does.
The serving engine keeps the first 30~s of audio for Gemma~4 12B.
Some requests remain too large and score 0: 25 of 16{,}000 questions for Clef and Clef-Flash, 8 of 17{,}500 for
Clef-Omni, and 172 of 17{,}500 for Jev-Omni with head v3, most of them whole AMI meetings.
The server of Xor-26B-A4B refuses requests with more than one video (500 questions).

\paragraph{Base models.}
The base MLLM of each System-1 model is the \texttt{base\_model} on its model card: Qwen3.8-27B for d3, Clef, and
OneJev-27B; Qwen3.5-9B for Clef-Flash and OneJev-9B; Qwen3.5 of the same size for OneJev-4B and OneJev-0.8B;
Qwen3-Omni-30B-A3B-Instruct for Clef-Omni; Gemma~4 12B for Jev-Omni; and Gemma~4 26B-A4B for Xor-26B-A4B.
Raya-Decision-v1 has no \texttt{base\_model}, but the text of its card names Qwen3.5-2B as its base.
We evaluate Qwen3.8-27B in its official FP8 release.
Precision changes the static accuracy little in our runs, even at 4 bits: Qwen2.5-Omni-7B reaches 55.5\%, 55.4\%, and
55.2\% in BF16, AWQ, and GPTQ, and DiffusionGemma-26B-A4B reaches 54.1\% and 54.2\% in BF16 and AWQ.

\section{Thinking Budgets}
\label{app:thinking}
Of the \nOpenMLLMs{} open MLLMs, \nThinkModels{} have an official thinking mode: Qwen3.8-27B, the five Qwen3.5 models,
Qwen3-Omni-30B-A3B-Thinking, Nemotron-3-Nano-Omni, Gemma~4 12B and 26B-A4B, DiffusionGemma-26B-A4B in BF16 and AWQ, and
MiniCPM-o 4.5.
The other \nNoThinkModels{} (Qwen3-VL-30B-A3B-Instruct, Qwen3-Omni-30B-A3B-Instruct, the four Qwen2.5-Omni models,
MiniCPM-o 2.6, and Phi-4-multimodal) run with thinking off only.
With thinking on, each question gets one thought, sampled with the model's own thinking settings and written up to
2{,}048 tokens, and the model does not see its budget.
For each budget $B \in \{0, 128, 256, 512, 1{,}024, 2{,}048\}$, we cut this thought after $B$ tokens (the whole
thought if it ended sooner), append the end-of-thought marker of the chat template, and read the label probabilities
right after it, as in Section~\ref{sec:setup}.
So the budgets differ only in where one sample is cut.
DiffusionGemma writes its thought with its own denoising sampler and is read with one read-only step after it.

Figure~\ref{fig:thinking} starts each curve at thinking off, the setting of Tables~\ref{tab:main}
and~\ref{tab:imgaud}.
Budget 0, the thinking prompt with an empty thought, generates no tokens either.
For the Qwen3.5 models, it is the same prompt as thinking off.
For Qwen3.8-27B, one more line sets the reasoning effort, and Gemma~4 and DiffusionGemma get a thinking flag.
At budget 0, these models are 2.0 to 6.4 points less accurate on images and video than with thinking off.
In closed loops, each decision gets one thought of up to 2{,}048 tokens, without cuts, and the episodes are paused as
in Table~\ref{tab:main}.
Three episodes ended without an outcome for reasons outside the model, two of Qwen3-Omni-30B-A3B-Thinking and one of
Nemotron-3-Nano-Omni, and are left out.

Table~\ref{tab:thinking} compares every model with thinking on and off.
Thinking raises the accuracy of all \nThinkModels{} models on images and of 12 on static video, and it raises the Brier
score of 10 on images and of 12 on static video.
In closed loops, all \nThinkModels{} gain in GameWorld, and four of them pass d3 there.
The thoughts are long: 331 to 1{,}286 tokens per question on video and 366 to 1{,}973 per decision in closed loops,
where up to 92\% of them reach the cap.

\begin{table*}[t]
\centering
\scriptsize
\setlength{\tabcolsep}{2.6pt}
\begin{adjustbox}{max width=\textwidth}
\begin{tabular}{@{}lrrrrrrrrrrrrrrr@{}}
\toprule
\multirow{3}{*}{Model}
& \multicolumn{8}{c}{Non-agentic} & \multicolumn{3}{c}{Agentic} \\
\cmidrule(lr){2-9} \cmidrule(lr){10-12}
& \multicolumn{2}{c}{Video static} & \multicolumn{2}{c}{AV} & \multicolumn{2}{c}{Image}
& \multicolumn{2}{c}{Audio} & \multicolumn{3}{c}{Closed loop}
& \multicolumn{4}{c}{Thought tokens} \\
\cmidrule(lr){2-3} \cmidrule(lr){4-5} \cmidrule(lr){6-7} \cmidrule(lr){8-9} \cmidrule(lr){10-12}
\cmidrule(lr){13-16}
& Acc. & Brier & Acc. & Brier & Acc. & Brier & Acc. & Brier & GameWorld & BALROG & VS-Bench & Video & Image & Audio
& Loop \\
\midrule
Qwen3.5-35B-A3B (Int4) & \textcolor{s1gain}{68.7} & \textcolor{s1loss}{0.569} & -- & -- & \textcolor{s1gain}{68.5} & \textcolor{s1loss}{0.545} & -- & -- & \textcolor{s1gain}{43.4} & \textcolor{s1gain}{13.3} & \textcolor{s1gain}{37.2} & 1{,}181 & 969 & -- & 1{,}959 \\
\hspace{0.5em}\textcolor{gray}{thinking off} & \textcolor{gray}{66.0} & \textcolor{gray}{0.427} & -- & -- & \textcolor{gray}{64.5} & \textcolor{gray}{0.461} & -- & -- & \textcolor{gray}{31.2} & \textcolor{gray}{7.0} & \textcolor{gray}{25.0} & & & & \\
Qwen3.5-9B & \textcolor{s1gain}{66.0} & \textcolor{s1loss}{0.607} & -- & -- & \textcolor{s1gain}{66.9} & \textcolor{s1loss}{0.560} & -- & -- & \textcolor{s1gain}{40.8} & \textcolor{s1gain}{9.5} & \textcolor{s1gain}{41.9} & 1{,}235 & 863 & -- & 1{,}973 \\
\hspace{0.5em}\textcolor{gray}{thinking off} & \textcolor{gray}{63.9} & \textcolor{gray}{0.457} & -- & -- & \textcolor{gray}{60.7} & \textcolor{gray}{0.492} & -- & -- & \textcolor{gray}{29.5} & \textcolor{gray}{3.7} & \textcolor{gray}{24.8} & & & & \\
Qwen3.8-27B (FP8) & \textcolor{s1gain}{67.0} & \textcolor{s1loss}{0.604} & -- & -- & \textcolor{s1gain}{70.6} & \textcolor{s1loss}{0.528} & -- & -- & \textcolor{s1gain}{49.3} & \textcolor{s1gain}{28.0} & \textcolor{s1gain}{43.5} & 1{,}099 & 407 & -- & 1{,}636 \\
\hspace{0.5em}\textcolor{gray}{thinking off} & \textcolor{gray}{62.5} & \textcolor{gray}{0.509} & -- & -- & \textcolor{gray}{64.8} & \textcolor{gray}{0.439} & -- & -- & \textcolor{gray}{34.6} & \textcolor{gray}{8.1} & \textcolor{gray}{19.9} & & & & \\
Qwen3.5-4B & \textcolor{s1gain}{64.6} & \textcolor{s1loss}{0.622} & -- & -- & \textcolor{s1gain}{64.1} & \textcolor{s1loss}{0.588} & -- & -- & \textcolor{s1gain}{39.4} & \textcolor{s1gain}{11.0} & \textcolor{s1gain}{38.2} & 1{,}285 & 951 & -- & 1{,}963 \\
\hspace{0.5em}\textcolor{gray}{thinking off} & \textcolor{gray}{62.0} & \textcolor{gray}{0.478} & -- & -- & \textcolor{gray}{57.3} & \textcolor{gray}{0.521} & -- & -- & \textcolor{gray}{31.8} & \textcolor{gray}{3.4} & \textcolor{gray}{27.6} & & & & \\
MiniCPM-o 4.5 & \textcolor{s1loss}{58.3} & \textcolor{s1loss}{0.759} & \textcolor{s1loss}{51.9} & \textcolor{s1loss}{0.786} & \textcolor{s1gain}{60.3} & \textcolor{s1loss}{0.710} & \textcolor{s1gain}{60.1} & \textcolor{s1loss}{0.712} & \textcolor{s1gain}{29.3} & \textcolor{s1gain}{9.4} & 24.7 & 331 & 384 & 287 & 366 \\
\hspace{0.5em}\textcolor{gray}{thinking off} & \textcolor{gray}{60.0} & \textcolor{gray}{0.543} & \textcolor{gray}{55.2} & \textcolor{gray}{0.742} & \textcolor{gray}{59.3} & \textcolor{gray}{0.563} & \textcolor{gray}{57.9} & \textcolor{gray}{0.548} & \textcolor{gray}{21.6} & \textcolor{gray}{2.4} & \textcolor{gray}{24.6} & & & & \\
Qwen3-Omni-30B-A3B (Thinking) & \textcolor{s1gain}{62.2} & \textcolor{s1loss}{0.727} & \textcolor{s1gain}{70.6} & \textcolor{s1gain}{0.541} & \textcolor{s1gain}{62.6} & 0.658 & \textcolor{s1gain}{63.6} & \textcolor{s1loss}{0.662} & \textcolor{s1gain}{36.1} & \textcolor{s1gain}{9.2} & \textcolor{s1gain}{26.3} & 365 & 303 & 327 & 1{,}183 \\
\hspace{0.5em}\textcolor{gray}{thinking off} & \textcolor{gray}{59.6} & \textcolor{gray}{0.617} & \textcolor{gray}{59.8} & \textcolor{gray}{0.628} & \textcolor{gray}{57.5} & \textcolor{gray}{0.666} & \textcolor{gray}{59.8} & \textcolor{gray}{0.570} & \textcolor{gray}{20.5} & \textcolor{gray}{2.5} & \textcolor{gray}{17.0} & & & & \\
Nemotron-3-Nano-Omni & \textcolor{s1gain}{60.9} & \textcolor{s1loss}{0.677} & \textcolor{s1loss}{63.7} & \textcolor{s1loss}{0.642} & \textcolor{s1gain}{61.6} & 0.638 & \textcolor{s1gain}{56.2} & \textcolor{s1loss}{0.805} & \textcolor{s1gain}{35.5} & \textcolor{s1gain}{6.4} & \textcolor{s1gain}{30.3} & 1{,}286 & 913 & 293 & 1{,}829 \\
\hspace{0.5em}\textcolor{gray}{thinking off} & \textcolor{gray}{58.7} & \textcolor{gray}{0.546} & \textcolor{gray}{65.5} & \textcolor{gray}{0.531} & \textcolor{gray}{54.6} & \textcolor{gray}{0.629} & \textcolor{gray}{54.4} & \textcolor{gray}{0.573} & \textcolor{gray}{18.8} & \textcolor{gray}{5.3} & \textcolor{gray}{22.0} & & & & \\
Gemma~4 26B-A4B & \textcolor{s1gain}{60.8} & \textcolor{s1gain}{0.756} & -- & -- & \textcolor{s1gain}{68.4} & \textcolor{s1gain}{0.591} & -- & -- & \textcolor{s1gain}{41.5} & \textcolor{s1loss}{8.5} & \textcolor{s1gain}{62.0} & 1{,}086 & 563 & -- & 1{,}717 \\
\hspace{0.5em}\textcolor{gray}{thinking off} & \textcolor{gray}{56.8} & \textcolor{gray}{0.777} & -- & -- & \textcolor{gray}{64.5} & \textcolor{gray}{0.629} & -- & -- & \textcolor{gray}{32.5} & \textcolor{gray}{9.7} & \textcolor{gray}{22.2} & & & & \\
DiffusionGemma-26B-A4B (AWQ) & \textcolor{s1gain}{57.2} & \textcolor{s1loss}{0.781} & -- & -- & \textcolor{s1gain}{63.6} & \textcolor{s1loss}{0.644} & -- & -- & \textcolor{s1gain}{38.0} & \textcolor{s1gain}{5.9} & \textcolor{s1gain}{26.6} & 741 & 276 & -- & 940 \\
\hspace{0.5em}\textcolor{gray}{thinking off} & \textcolor{gray}{54.2} & \textcolor{gray}{0.663} & -- & -- & \textcolor{gray}{60.7} & \textcolor{gray}{0.579} & -- & -- & \textcolor{gray}{25.6} & \textcolor{gray}{4.1} & \textcolor{gray}{9.2} & & & & \\
DiffusionGemma-26B-A4B & \textcolor{s1gain}{58.4} & \textcolor{s1loss}{0.776} & -- & -- & \textcolor{s1gain}{65.4} & \textcolor{s1loss}{0.622} & -- & -- & \textcolor{s1gain}{40.8} & 7.0 & \textcolor{s1gain}{49.7} & 640 & 246 & -- & 740 \\
\hspace{0.5em}\textcolor{gray}{thinking off} & \textcolor{gray}{54.1} & \textcolor{gray}{0.651} & -- & -- & \textcolor{gray}{61.9} & \textcolor{gray}{0.543} & -- & -- & \textcolor{gray}{25.9} & \textcolor{gray}{6.2} & \textcolor{gray}{11.7} & & & & \\
Qwen3.5-2B & \textcolor{s1gain}{58.7} & \textcolor{s1loss}{0.716} & -- & -- & \textcolor{s1gain}{54.9} & \textcolor{s1loss}{0.745} & -- & -- & \textcolor{s1gain}{31.1} & \textcolor{s1gain}{3.0} & 15.5 & 1{,}217 & 945 & -- & 1{,}973 \\
\hspace{0.5em}\textcolor{gray}{thinking off} & \textcolor{gray}{53.0} & \textcolor{gray}{0.555} & -- & -- & \textcolor{gray}{46.6} & \textcolor{gray}{0.647} & -- & -- & \textcolor{gray}{13.6} & \textcolor{gray}{1.3} & \textcolor{gray}{15.9} & & & & \\
Gemma~4 12B & \textcolor{s1gain}{56.3} & \textcolor{s1loss}{0.837} & \textcolor{s1gain}{61.3} & \textcolor{s1gain}{0.730} & \textcolor{s1gain}{66.5} & \textcolor{s1gain}{0.627} & \textcolor{s1loss}{51.2} & \textcolor{s1loss}{0.887} & \textcolor{s1gain}{42.3} & \textcolor{s1gain}{11.4} & \textcolor{s1gain}{43.2} & 960 & 498 & 304 & 1{,}635 \\
\hspace{0.5em}\textcolor{gray}{thinking off} & \textcolor{gray}{51.8} & \textcolor{gray}{0.790} & \textcolor{gray}{46.4} & \textcolor{gray}{0.964} & \textcolor{gray}{61.6} & \textcolor{gray}{0.647} & \textcolor{gray}{53.0} & \textcolor{gray}{0.730} & \textcolor{gray}{25.1} & \textcolor{gray}{7.5} & \textcolor{gray}{40.4} & & & & \\
Qwen3.5-0.8B & \textcolor{s1gain}{52.0} & \textcolor{s1loss}{0.844} & -- & -- & \textcolor{s1gain}{48.0} & \textcolor{s1loss}{0.881} & -- & -- & \textcolor{s1gain}{31.2} & \textcolor{s1gain}{4.5} & \textcolor{s1gain}{24.0} & 1{,}103 & 895 & -- & 1{,}868 \\
\hspace{0.5em}\textcolor{gray}{thinking off} & \textcolor{gray}{47.9} & \textcolor{gray}{0.632} & -- & -- & \textcolor{gray}{38.8} & \textcolor{gray}{0.695} & -- & -- & \textcolor{gray}{18.5} & \textcolor{gray}{0.0} & \textcolor{gray}{9.6} & & & & \\
\bottomrule
\end{tabular}
\end{adjustbox}
\caption{\textbf{The \nThinkModels{} MLLMs with a thinking mode, with thinking on.} Each model's row gives
its scores with one thought of at most 2{,}048 tokens per question (per decision in closed loops), and the gray
row under it its scores with thinking off, from Tables~\ref{tab:main} and~\ref{tab:imgaud}. Green: thinking on is
better by at least 1 point (0.01 in Brier score), orange: worse by at least as much. The columns are those of
Table~\ref{tab:pairs}. VS-Bench is normalized per game with the range of Table~\ref{tab:main}, which holds the
models with thinking off. Thought tokens: the mean per question, and per decision in closed loops (video: the static and
audio-visual questions). --: the model does not take audio.}
\label{tab:thinking}
\end{table*}

\section{Domains and Sources}
\label{app:sources}

Tables~\ref{tab:sources_img}, \ref{tab:sources}, and~\ref{tab:sources_aud} list the source of every image, video, and
audio task with its license and the tier of its instructions.
The question counts count requests: every video request asks one question, while 10 image and 3 audio sources ask
several per request, for example one per privacy attribute in VISPR.
Tables~\ref{tab:leaves} and~\ref{tab:leaves_mllm} give the accuracy of each model on each video task.

\begin{table*}[t]
\centering
\scriptsize
\setlength{\tabcolsep}{2.5pt}
\begin{adjustbox}{max width=\textwidth}
\begin{tabular}{@{}lrrrrrrrrrrrrrr@{}}
\toprule
Task & Src. & Q. & \rotatebox{90}{d3} & \rotatebox{90}{Clef} & \rotatebox{90}{OneJev-27B} & \rotatebox{90}{OneJev-9B} & \rotatebox{90}{Clef-Flash} & \rotatebox{90}{OneJev-4B} & \rotatebox{90}{Clef-Omni} & \rotatebox{90}{Xor-26B-A4B} & \rotatebox{90}{OneJev-0.8B} & \rotatebox{90}{Jev-Omni (head v3)} & \rotatebox{90}{Jev-Omni} & \rotatebox{90}{Raya-Decision-v1} \\
\midrule
\multicolumn{15}{@{}l}{\textbf{Short video}} \\
basic/action\_candidates & 1 & 500 & 95.4 & \textbf{96.6} & 95.0 & 96.0 & \textbf{96.6} & 94.8 & 95.8 & 94.4 & 88.0 & 92.2 & 91.2 & 87.8 \\
basic/action\_category & 1 & 500 & \textbf{88.6} & 86.2 & 82.8 & 86.2 & 87.4 & 87.2 & 86.8 & 81.2 & 68.2 & 30.4 & 31.0 & 10.6 \\
basic/repetition & 2 & 500 & \textbf{25.8} & 25.6 & 16.4 & 14.8 & 19.6 & 13.8 & 14.0 & 18.0 & 10.4 & 14.4 & 18.2 & 13.8 \\
basic/state\_change & 2 & 500 & 73.0 & 70.9 & 75.9 & \textbf{79.5} & 72.4 & 77.3 & 71.7 & 66.3 & 59.7 & 63.5 & 67.2 & 54.9 \\
basic/temporal & 1 & 500 & \textbf{96.6} & 95.2 & 92.6 & 90.4 & 82.4 & 85.4 & 68.4 & 59.4 & 63.4 & 78.4 & 74.2 & 55.6 \\
batch/quality & 1 & 500 & 60.0 & 60.0 & 56.0 & 53.4 & 57.6 & 62.8 & 57.8 & 38.2 & \textbf{66.6} & 39.4 & 50.8 & 58.6 \\
bench/mvbench & 1 & 500 & \textbf{75.2} & 73.8 & 68.6 & 67.0 & 65.6 & 63.8 & 65.2 & 55.6 & 52.6 & 57.8 & 56.6 & 53.8 \\
bench/nextqa & 1 & 500 & 85.4 & 84.4 & 84.0 & 84.2 & \textbf{86.0} & 82.4 & 82.4 & 73.8 & 72.6 & 71.0 & 70.0 & 69.6 \\
bench/tempcompass & 1 & 500 & 83.6 & \textbf{83.8} & 82.6 & 79.0 & 80.4 & 76.4 & 72.8 & 71.4 & 67.0 & 70.6 & 69.6 & 56.6 \\
embodied/feasibility & 1 & 500 & 62.8 & 58.8 & 61.2 & \textbf{63.8} & 63.4 & 60.6 & 60.8 & 49.2 & 56.4 & 52.2 & 48.2 & 48.4 \\
embodied/procedure & 1 & 500 & 51.0 & 51.4 & \textbf{53.2} & 51.8 & 51.0 & 50.8 & 50.0 & 52.8 & 50.0 & 50.8 & 50.8 & 49.6 \\
embodied/robot\_success & 1 & 500 & 60.0 & \textbf{63.6} & 56.2 & 59.8 & 56.4 & 57.8 & 50.0 & 52.0 & 49.2 & 54.2 & 48.6 & 49.6 \\
fore/next\_action & 1 & 500 & \textbf{70.6} & 60.4 & 51.8 & 46.2 & 55.6 & 46.6 & 54.0 & 63.2 & 36.2 & 56.2 & 51.2 & 31.8 \\
fore/pedestrian & 2 & 500 & 55.4 & 52.9 & 55.0 & 59.8 & 50.4 & \textbf{61.0} & 49.1 & 48.9 & 55.5 & 49.2 & 50.5 & 55.5 \\
game/physics\_bug & 1 & 500 & 62.4 & \textbf{63.8} & 62.8 & 56.2 & 62.6 & 57.8 & 52.4 & 52.0 & 43.6 & 45.8 & 46.2 & 37.8 \\
judge/hallucination & 1 & 500 & \textbf{85.2} & 81.8 & 83.8 & 82.0 & 82.6 & 79.6 & 80.2 & 79.6 & 64.2 & 73.6 & 76.4 & 69.6 \\
judge/t2v & 1 & 500 & 39.8 & 41.0 & 39.0 & 40.4 & 38.8 & 41.6 & 32.0 & 30.6 & \textbf{45.8} & 39.4 & 37.2 & 35.8 \\
know/causal\_qa & 1 & 500 & \textbf{69.0} & 66.0 & 64.2 & 68.2 & 61.4 & 59.2 & 56.4 & 45.8 & 45.2 & 46.0 & 44.0 & 40.8 \\
know/physics & 1 & 500 & \textbf{55.6} & 52.2 & 50.8 & 53.4 & 52.8 & 55.2 & 51.2 & 51.0 & 53.0 & 51.0 & 50.6 & 51.8 \\
pro/driving\_risk & 1 & 500 & 51.8 & 54.4 & 53.2 & 54.0 & 49.6 & 54.4 & 49.8 & 50.2 & 50.0 & 50.6 & \textbf{54.6} & 50.0 \\
pro/foul & 1 & 500 & \textbf{48.6} & 47.0 & 47.4 & 46.0 & 40.0 & 48.0 & 41.4 & 0.0 & 43.4 & 27.2 & 26.8 & 41.2 \\
ret/text\_match & 1 & 500 & \textbf{98.6} & \textbf{98.6} & 97.8 & 97.6 & 98.4 & 97.4 & 96.8 & 95.6 & 92.2 & 95.6 & 94.6 & 86.0 \\
safety/violence & 2 & 500 & 88.0 & 92.0 & \textbf{93.1} & 92.6 & 86.3 & 91.9 & 83.1 & 77.8 & 92.8 & 81.3 & 84.5 & 78.6 \\
\multicolumn{15}{@{}l}{\textbf{Long video}} \\
bench/egoschema & 1 & 500 & 77.4 & 77.8 & \textbf{79.0} & 75.0 & 70.8 & 73.4 & 71.0 & 66.4 & 48.4 & 54.4 & 53.4 & 42.6 \\
bench/vsi\_bench & 1 & 500 & 59.4 & 67.8 & 69.6 & 67.4 & \textbf{70.8} & 64.2 & 60.0 & 45.2 & 39.8 & 36.4 & 31.6 & 29.8 \\
dialog/meeting\_decision & 1 & 500 & -- & -- & -- & -- & -- & -- & \textbf{71.0} & -- & -- & 31.8 & 52.2 & -- \\
embodied/ego\_plan & 1 & 500 & 44.0 & \textbf{49.0} & 48.8 & 44.6 & 38.2 & 34.6 & 37.4 & 32.4 & 32.2 & 30.8 & 29.8 & 22.6 \\
judge/meeting\_summary & 1 & 500 & -- & -- & -- & -- & -- & -- & \textbf{67.8} & -- & -- & 60.0 & 58.0 & -- \\
know/lecture & 1 & 500 & -- & -- & -- & -- & -- & -- & \textbf{69.8} & -- & -- & 28.4 & 26.4 & -- \\
know/long\_qa & 1 & 500 & 60.8 & \textbf{63.0} & 57.0 & 54.2 & 61.6 & 48.8 & 59.0 & 53.2 & 35.0 & 50.4 & 50.8 & 32.8 \\
know/needle & 1 & 500 & 79.0 & 83.0 & \textbf{94.2} & 82.2 & 74.4 & 77.6 & 64.2 & 52.6 & 53.2 & 43.4 & 42.2 & 30.2 \\
pro/surveillance & 1 & 500 & 71.8 & 76.2 & \textbf{77.0} & 71.2 & 75.6 & 71.8 & 71.0 & 61.0 & 62.2 & 67.8 & 66.6 & 38.6 \\
ret/moment & 3 & 500 & \textbf{61.4} & 61.3 & 57.8 & 57.1 & 58.0 & 44.2 & 60.7 & 43.8 & 29.0 & 39.7 & 40.2 & 29.3 \\
route/escalation & 1 & 500 & \textbf{64.6} & 51.6 & 56.4 & 50.4 & 50.8 & 51.6 & 49.8 & 50.2 & 50.0 & 51.8 & 50.2 & 47.8 \\
route/segment & 1 & 500 & 40.2 & 39.2 & 40.8 & \textbf{43.6} & 28.6 & 41.4 & 30.4 & 38.4 & 38.0 & 36.8 & 33.6 & 13.6 \\
\bottomrule
\end{tabular}
\end{adjustbox}
\caption{\textbf{Accuracy (\%) per video task: System-1 models.} Src.:
sources; Q.: questions; --: the model does not take audio. A task with several sources averages their accuracies. Bold:
the best model of the row in this table.}
\label{tab:leaves}
\end{table*}

\begin{table*}[t]
\centering
\scriptsize
\setlength{\tabcolsep}{2.5pt}
\begin{adjustbox}{max width=\textwidth}
\begin{tabular}{@{}lrrrrrrrrrrrrrrrrrrrrrrr@{}}
\toprule
Task & Src. & Q. & \rotatebox{90}{Qwen3.5-35B-A3B (Int4)} & \rotatebox{90}{Qwen3.5-9B} & \rotatebox{90}{Qwen3-VL-30B-A3B} & \rotatebox{90}{Qwen3.8-27B (FP8)} & \rotatebox{90}{Qwen3.5-4B} & \rotatebox{90}{MiniCPM-o 4.5} & \rotatebox{90}{Qwen3-Omni-30B-A3B (Thinking)} & \rotatebox{90}{Nemotron-3-Nano-Omni} & \rotatebox{90}{Qwen3-Omni-30B-A3B} & \rotatebox{90}{Gemma~4 26B-A4B} & \rotatebox{90}{Qwen2.5-Omni-7B} & \rotatebox{90}{Qwen2.5-Omni-7B (AWQ)} & \rotatebox{90}{Qwen2.5-Omni-7B (GPTQ)} & \rotatebox{90}{DiffusionGemma-26B-A4B (AWQ)} & \rotatebox{90}{DiffusionGemma-26B-A4B} & \rotatebox{90}{Qwen2.5-Omni-3B} & \rotatebox{90}{MiniCPM-o 2.6} & \rotatebox{90}{Qwen3.5-2B} & \rotatebox{90}{Gemma~4 12B} & \rotatebox{90}{Phi-4-multimodal} & \rotatebox{90}{Qwen3.5-0.8B} \\
\midrule
\multicolumn{24}{@{}l}{\textbf{Short video}} \\
basic/action\_candidates & 1 & 500 & 96.8 & 96.6 & \textbf{97.4} & 95.6 & 96.8 & 94.4 & 97.0 & 97.0 & 97.2 & 95.0 & 95.4 & 95.8 & 96.2 & 94.0 & 93.4 & 95.6 & 93.4 & 95.8 & 92.4 & 88.2 & 92.4 \\
basic/action\_category & 1 & 500 & \textbf{89.8} & 85.8 & 85.8 & 84.6 & 86.2 & 76.6 & 83.6 & 73.4 & 85.2 & 82.8 & 43.6 & 45.4 & 41.0 & 81.0 & 81.2 & 30.4 & 37.0 & 79.0 & 66.4 & 48.4 & 49.6 \\
basic/repetition & 2 & 500 & 26.8 & 26.6 & 22.6 & \textbf{28.6} & 22.6 & 15.2 & 14.8 & 20.4 & 13.2 & 11.0 & 16.2 & 15.4 & 10.4 & 20.4 & 18.4 & 4.8 & 20.8 & 11.0 & 15.4 & 15.8 & 11.4 \\
basic/state\_change & 2 & 500 & 73.6 & 74.5 & 74.7 & 74.3 & 76.2 & \textbf{76.4} & 76.0 & 69.6 & 70.6 & 71.1 & 69.2 & 69.8 & 66.9 & 67.3 & 66.0 & 63.4 & 70.2 & 56.2 & 63.7 & 57.8 & 57.8 \\
basic/temporal & 1 & 500 & 87.4 & \textbf{88.6} & 79.0 & 86.6 & 81.4 & 79.0 & 69.0 & 76.0 & 64.6 & 66.0 & 70.8 & 72.2 & 71.8 & 59.0 & 61.4 & 52.8 & 64.4 & 60.0 & 61.2 & 53.0 & 49.2 \\
batch/quality & 1 & 500 & 68.0 & \textbf{68.8} & 49.8 & 48.6 & 56.2 & 42.2 & 66.8 & 35.4 & 21.8 & 41.6 & 26.4 & 25.6 & 46.8 & 51.6 & 46.6 & 54.4 & 38.0 & 27.4 & 33.0 & 59.6 & 0.2 \\
bench/mvbench & 1 & 500 & \textbf{73.0} & 72.0 & 72.4 & 71.8 & 68.2 & 63.2 & 66.8 & 70.8 & 65.6 & 56.0 & 71.8 & 71.6 & 71.0 & 52.4 & 51.6 & 66.4 & 61.6 & 62.6 & 51.4 & 51.0 & 56.0 \\
bench/nextqa & 1 & 500 & \textbf{87.0} & 84.0 & 84.8 & 83.0 & 81.4 & 81.8 & 83.8 & 83.2 & 80.8 & 75.0 & 81.0 & 80.2 & 80.4 & 72.4 & 71.6 & 79.2 & 78.4 & 77.2 & 68.2 & 68.8 & 70.0 \\
bench/tempcompass & 1 & 500 & 80.6 & 75.4 & 76.8 & \textbf{82.2} & 76.8 & 69.0 & 72.6 & 72.8 & 72.4 & 70.2 & 69.6 & 69.2 & 70.6 & 68.6 & 69.6 & 66.4 & 68.2 & 65.8 & 67.4 & 62.6 & 61.0 \\
embodied/feasibility & 1 & 500 & 65.6 & 61.2 & \textbf{66.4} & 60.4 & 62.0 & 57.8 & 61.8 & 53.0 & 63.6 & 52.6 & 64.4 & 65.0 & 64.0 & 50.6 & 48.6 & 62.6 & 52.2 & 53.4 & 42.8 & 44.6 & 54.2 \\
embodied/procedure & 1 & 500 & 50.4 & 50.2 & 50.4 & 50.0 & 50.0 & \textbf{52.2} & 49.2 & 50.0 & 50.2 & 49.6 & 50.0 & 50.0 & 50.0 & 49.8 & 50.0 & 50.6 & 50.8 & 49.8 & 50.0 & 50.8 & 50.0 \\
embodied/robot\_success & 1 & 500 & \textbf{60.6} & 54.0 & 54.0 & 50.6 & 60.4 & 52.6 & 56.8 & 51.6 & 54.8 & 54.4 & 53.8 & 53.6 & 51.6 & 50.6 & 51.4 & 51.2 & 49.6 & 50.2 & 51.0 & 50.0 & 50.0 \\
fore/next\_action & 1 & 500 & \textbf{67.0} & 53.8 & 64.4 & 65.4 & 59.2 & 59.0 & 59.6 & 64.0 & 49.2 & 59.0 & 49.2 & 49.8 & 48.2 & 45.4 & 48.2 & 48.8 & 47.6 & 43.2 & 59.6 & 57.0 & 47.8 \\
fore/pedestrian & 2 & 500 & 55.4 & 49.4 & 51.3 & 49.4 & 54.5 & 50.1 & 47.9 & 54.1 & 48.4 & 48.5 & 55.1 & 56.2 & 54.4 & 49.2 & 48.4 & 51.5 & \textbf{56.5} & 51.8 & 48.4 & 51.6 & 47.5 \\
game/physics\_bug & 1 & 500 & 60.0 & 57.8 & 57.8 & \textbf{60.8} & 57.2 & 51.0 & 54.8 & 49.8 & 55.2 & 53.4 & 50.0 & 50.4 & 49.6 & 47.6 & 48.0 & 46.2 & 51.2 & 49.8 & 43.6 & 40.6 & 42.0 \\
judge/hallucination & 1 & 500 & \textbf{84.2} & 82.2 & 83.0 & 77.4 & 82.2 & 82.4 & 75.4 & 82.0 & 82.0 & 78.2 & 82.0 & 81.0 & 81.2 & 80.6 & 78.0 & 76.4 & 78.2 & 63.8 & 76.4 & 76.8 & 67.8 \\
judge/t2v & 1 & 500 & 15.4 & 23.4 & 17.0 & 21.4 & 11.6 & 27.6 & \textbf{32.4} & 9.8 & 29.8 & 20.6 & 17.0 & 18.0 & 20.0 & 24.8 & 23.2 & 16.2 & 18.8 & 32.2 & 17.0 & 8.6 & 18.4 \\
know/causal\_qa & 1 & 500 & 65.6 & \textbf{67.0} & 65.8 & 65.0 & 65.4 & 56.4 & 64.0 & 66.0 & 57.2 & 49.4 & 55.8 & 51.8 & 52.6 & 44.8 & 45.6 & 50.6 & 50.2 & 46.2 & 44.6 & 40.4 & 39.8 \\
know/physics & 1 & 500 & 53.4 & 51.8 & 49.4 & 52.2 & 50.8 & 53.2 & 52.8 & 50.4 & \textbf{55.2} & 49.8 & 50.6 & 52.8 & 51.8 & 46.2 & 50.4 & 53.0 & 53.6 & 51.4 & 51.4 & 51.0 & 52.4 \\
pro/driving\_risk & 1 & 500 & 53.2 & 53.6 & 57.4 & 51.4 & 50.0 & 57.4 & 50.2 & 51.2 & 50.0 & 51.0 & 50.6 & 52.6 & 52.8 & 50.0 & 50.0 & \textbf{60.8} & 49.2 & 51.0 & 50.0 & 50.0 & 50.0 \\
pro/foul & 1 & 500 & 43.6 & 45.2 & 40.6 & \textbf{48.2} & 44.6 & 47.0 & 42.4 & 48.0 & 38.2 & 41.6 & 42.4 & 42.2 & 45.2 & 29.2 & 28.4 & 44.8 & 38.8 & 37.4 & 29.2 & 24.4 & 36.4 \\
ret/text\_match & 1 & 500 & \textbf{97.8} & 97.6 & 96.8 & 97.0 & 97.6 & 96.4 & 97.0 & 97.0 & 97.6 & 96.2 & \textbf{97.8} & 97.6 & \textbf{97.8} & 96.8 & 96.6 & 97.0 & 96.4 & 95.0 & 93.0 & 94.0 & 92.6 \\
safety/violence & 2 & 500 & 94.5 & 90.0 & 90.1 & 90.7 & 92.1 & 89.7 & 88.1 & \textbf{95.9} & 85.1 & 85.5 & 94.0 & 93.3 & 92.8 & 83.3 & 82.7 & 91.3 & 89.4 & 91.0 & 73.3 & 89.7 & 77.1 \\
\multicolumn{24}{@{}l}{\textbf{Long video}} \\
bench/egoschema & 1 & 500 & \textbf{77.0} & 74.2 & 73.0 & 76.0 & 72.2 & 65.0 & 68.2 & 68.4 & 65.4 & 64.6 & 70.2 & 68.4 & 68.0 & 61.4 & 61.8 & 63.6 & 56.6 & 56.2 & 53.4 & 51.4 & 46.0 \\
bench/vsi\_bench & 1 & 500 & 68.8 & \textbf{70.4} & 64.4 & 53.2 & 54.0 & 49.2 & 47.0 & 44.6 & 54.4 & 43.4 & 34.4 & 33.4 & 33.2 & 39.6 & 41.2 & 31.0 & 37.8 & 48.2 & 36.0 & 32.6 & 41.6 \\
dialog/meeting\_decision & 1 & 500 & -- & -- & -- & -- & -- & 51.4 & 50.8 & 61.8 & 53.2 & -- & 55.6 & 56.8 & 55.4 & -- & -- & 56.0 & 47.8 & -- & 30.6 & \textbf{68.4} & -- \\
embodied/ego\_plan & 1 & 500 & 45.4 & 38.0 & 45.6 & \textbf{48.0} & 39.4 & 29.4 & 33.8 & 28.2 & 36.8 & 35.2 & 27.8 & 27.8 & 26.4 & 26.8 & 29.4 & 29.0 & 27.6 & 33.0 & 32.6 & 29.4 & 27.2 \\
judge/meeting\_summary & 1 & 500 & -- & -- & -- & -- & -- & 67.8 & 68.4 & \textbf{71.6} & 64.4 & -- & 65.0 & 66.0 & 66.0 & -- & -- & 64.4 & 65.8 & -- & 52.4 & 58.8 & -- \\
know/lecture & 1 & 500 & -- & -- & -- & -- & -- & 46.4 & 60.2 & \textbf{63.2} & 48.4 & -- & 52.2 & 49.8 & 47.8 & -- & -- & 45.0 & 42.6 & -- & 56.2 & 49.2 & -- \\
know/long\_qa & 1 & 500 & \textbf{61.8} & 57.6 & 58.4 & 60.0 & 51.2 & 56.6 & 53.0 & 54.8 & 53.2 & 50.4 & 48.6 & 46.4 & 42.6 & 46.6 & 45.2 & 46.4 & 49.2 & 47.0 & 47.2 & 41.8 & 37.0 \\
know/needle & 1 & 500 & \textbf{84.6} & 72.0 & 74.4 & 64.4 & 73.6 & 71.6 & 50.2 & 66.4 & 64.2 & 60.0 & 45.8 & 45.2 & 42.8 & 48.6 & 47.8 & 35.2 & 38.8 & 45.6 & 50.6 & 40.2 & 43.0 \\
pro/surveillance & 1 & 500 & \textbf{75.6} & 72.8 & 72.6 & 54.8 & 71.4 & 73.8 & 53.0 & 56.4 & 69.8 & 69.0 & 70.6 & 69.6 & 65.2 & 69.4 & 69.4 & 48.6 & 51.8 & 40.2 & 63.4 & 38.8 & 41.4 \\
ret/moment & 3 & 500 & 58.7 & 59.8 & 52.3 & \textbf{60.0} & 54.3 & 52.9 & 51.9 & 50.7 & 45.1 & 49.3 & 47.9 & 47.5 & 45.7 & 41.2 & 41.5 & 49.3 & 49.6 & 41.2 & 37.3 & 36.3 & 41.6 \\
route/escalation & 1 & 500 & 51.4 & 50.6 & 47.2 & 51.0 & 49.0 & \textbf{54.6} & 48.0 & 52.0 & 48.0 & 49.4 & 48.2 & 48.8 & 50.2 & 48.6 & 48.6 & 53.8 & 47.8 & 49.6 & 51.0 & 47.8 & 49.6 \\
route/segment & 1 & 500 & 38.4 & \textbf{39.4} & 37.2 & 38.8 & 35.2 & 37.6 & 37.8 & 34.0 & 38.6 & 36.8 & 25.2 & 27.0 & 24.6 & 37.6 & 38.4 & 39.2 & 36.6 & 35.2 & 36.0 & 36.8 & 32.0 \\
\bottomrule
\end{tabular}
\end{adjustbox}
\caption{\textbf{Accuracy (\%) per video task: Open MLLMs, read at the answer token.} Src.:
sources; Q.: questions; --: the model does not take audio. A task with several sources averages their accuracies. Bold:
the best model of the row in this table.}
\label{tab:leaves_mllm}
\end{table*}

\begin{table*}[p]
\centering
\scriptsize
\setlength{\tabcolsep}{3pt}
\begin{tabular}{@{}>{\raggedright\arraybackslash}p{5.2cm}>{\raggedright\arraybackslash}p{1.1cm}%
>{\raggedright\arraybackslash}p{8.6cm}cr@{}}
\toprule
Task and source & Ref. & License & Tier & Questions \\
\midrule
img.basic/\allowbreak{}custom\_categories/\allowbreak{}cub & \cite{WahCUB_200_2011} & non-commercial research and education; CaltechDATA record lists CC BY & B & 500 \\
img.basic/\allowbreak{}fine\_grained/\allowbreak{}cub & \cite{WahCUB_200_2011} & non-commercial research and education; CaltechDATA record lists CC BY & B & 500 \\
img.basic/\allowbreak{}visual\_entailment/\allowbreak{}esnlive & \cite{Kayser_2021_ICCV} & no license stated, SNLI labels CC BY-SA 4.0, Flickr30k images for non-commercial research & B & 500 \\
img.basic/\allowbreak{}visual\_entailment/\allowbreak{}nlvr2 & \cite{suhr-etal-2019-corpus} & CC BY 4.0 annotations; web images & B & 500 \\
img.batch/\allowbreak{}curation/\allowbreak{}koniq10k & \cite{koniq10k} & no license stated, Flickr images under their own Creative Commons licenses & B & 500 \\
img.batch/\allowbreak{}curation/\allowbreak{}seetrue & \cite{yarom2023you} & CC BY 4.0, test-only use & B & 500 \\
img.code/\allowbreak{}render\_match/\allowbreak{}matplotlib\_render & -- & CC BY 4.0 (generated) & C & 500 \\
img.embodied/\allowbreak{}eb\_alfred\_visual/\allowbreak{}eb\_alfred & \cite{yang2025embodiedbench} & not stated on the dataset card & B & 500 \\
img.game/\allowbreak{}atari\_head/\allowbreak{}atari\_head & \cite{zhang2019atarihead,zhang2018agil} & CC BY 4.0 & B & 500 \\
img.game/\allowbreak{}board\_twin/\allowbreak{}lichess\_puzzles & \cite{lichess2024puzzles} & CC0 & B & 500 \\
img.game/\allowbreak{}glitch/\allowbreak{}glitchbench & \cite{Taesiri_2024_CVPR} & MIT (card); screenshots from Reddit & B & 249 \\
img.game/\allowbreak{}glitch/\allowbreak{}videogameqa\_bench & \cite{NEURIPS2025_db2cf8f4} & CC BY 4.0 & A & 500 \\
img.judge/\allowbreak{}agent\_success/\allowbreak{}agentrewardbench & \cite{lu2025agentrewardbench} & dataset terms of use (research) & B & 500 \\
img.judge/\allowbreak{}hallucination/\allowbreak{}hallusionbench & \cite{Guan_2024_CVPR} & BSD-3-Clause & A & 499 \\
img.judge/\allowbreak{}preference/\allowbreak{}vl\_rewardbench & \cite{VLRewardBench} & MIT & B & 500 \\
img.judge/\allowbreak{}t2i\_alignment/\allowbreak{}genai\_bench & \cite{li2024genai} & Apache-2.0 & B & 387 \\
img.know/\allowbreak{}clock\_calendar/\allowbreak{}rendered & -- & CC BY 4.0 (generated) & C & 1,000 \\
img.know/\allowbreak{}math\_visual/\allowbreak{}mathvista & \cite{lu2024mathvista} & CC BY-SA 4.0 & A & 495 \\
img.know/\allowbreak{}perception/\allowbreak{}blink & \cite{fu2024blink} & Apache-2.0 & A & 500 \\
img.know/\allowbreak{}perception/\allowbreak{}countbench & \cite{paiss2023countclip} & unspecified; web images & B & 479 \\
img.know/\allowbreak{}subject\_mcq/\allowbreak{}mmmu\_pro & \cite{yue2024mmmu} & Apache-2.0 & A & 500 \\
img.know/\allowbreak{}visual\_rules/\allowbreak{}bongard\_openworld & \cite{wu2024bongardopenworld} & CC BY-NC-SA 4.0; images crawled from the web & B & 400 \\
img.pro/\allowbreak{}damage/\allowbreak{}resisc45 & \cite{Cheng_2017} & CC BY-NC 4.0 & B & 500 \\
img.pro/\allowbreak{}damage/\allowbreak{}xbd & \cite{gupta2019xbddatasetassessingbuilding} & CC BY-NC-SA 4.0 & B & 500 \\
img.pro/\allowbreak{}defect/\allowbreak{}mvtec\_ad & \cite{Bergmann_2019_CVPR,Bergmann2021} & CC BY-NC-SA 4.0 & B & 500 \\
img.pro/\allowbreak{}medical/\allowbreak{}slake & \cite{Liu_2021} & CC BY 4.0 & A & 850 \\
img.pro/\allowbreak{}phishing\_page/\allowbreak{}phishpedia & \cite{lin2021phishpedia} & CC0 1.0 (repository license) & B & 500 \\
img.ret/\allowbreak{}compositional/\allowbreak{}winoground & \cite{thrush_and_ross2022winoground} & Winoground license (Meta): research only, no distributing its images or metadata & B & 500 \\
img.ret/\allowbreak{}same\_item/\allowbreak{}stanford\_online\_products & \cite{songCVPR16} & MIT (repository license) & B & 500 \\
img.route/\allowbreak{}abstain/\allowbreak{}mmmu\_pro\_masked & \cite{yue2024mmmu} & Apache-2.0 (MMMU-Pro) & C & 500 \\
img.route/\allowbreak{}document\_routing/\allowbreak{}rvl\_cdip & \cite{harley2015icdar} & research use (IIT-CDIP) & B & 500 \\
img.route/\allowbreak{}model\_correctness/\allowbreak{}mmmu\_pro\_small\_vlm & \cite{yue2024mmmu} & Apache-2.0 (MMMU-Pro) & C & 500 \\
img.route/\allowbreak{}quality\_gate/\allowbreak{}vizwiz\_quality & \cite{Chiu_2020_CVPR} & CC BY 4.0 & B & 500 \\
img.route/\allowbreak{}quality\_gate/\allowbreak{}vizwiz\_vqa & \cite{gurari2018vizwiz,Gurari_2019_CVPR} & CC BY 4.0 & B & 500 \\
img.safety/\allowbreak{}hateful\_meme/\allowbreak{}hateful\_memes & \cite{Kiela:2020hatefulmemes} & Hateful Memes Dataset License (non-commercial research) & B & 500 \\
img.safety/\allowbreak{}injection/\allowbreak{}figstep & \cite{gong2025figstep} & MIT & C & 295 \\
img.safety/\allowbreak{}policy/\allowbreak{}unsafebench & \cite{QSWBZZ24} & UnsafeBench data use agreement (research, education, responsible commercial use) & B & 500 \\
img.safety/\allowbreak{}privacy/\allowbreak{}vispr & \cite{orekondy17iccv} & CC BY-NC 4.0 & B & 500 \\
img.safety/\allowbreak{}synthetic/\allowbreak{}genimage\_midjourney & \cite{zhu2023genimage} & CC BY-NC-SA 4.0, ImageNet images under the ImageNet terms & B & 500 \\
img.social/\allowbreak{}art\_emotion/\allowbreak{}artemis & \cite{youssef2022artemis2} & ArtEmis terms of use (non-commercial research and education) & B & 500 \\
img.social/\allowbreak{}humor/\allowbreak{}newyorker & \cite{newyorkernextmldataset,zhang2024humor} & CC BY-NC 4.0; cartoons copyrighted & B & 500 \\
img.std/\allowbreak{}aokvqa/\allowbreak{}aokvqa & \cite{Schwenk_2022} & Apache-2.0 (annotations); COCO images & A & 498 \\
img.std/\allowbreak{}gqa/\allowbreak{}gqa & \cite{hudson2018gqa} & No license stated; images and scene graphs from Visual Genome (CC BY 4.0) & B & 461 \\
img.std/\allowbreak{}mmmu/\allowbreak{}mmmu & \cite{yue2023mmmu} & Apache-2.0 (release); images from the sources it cites & A & 497 \\
img.std/\allowbreak{}naturalbench/\allowbreak{}naturalbench & \cite{naturalbench} & Apache-2.0 & A & 498 \\
img.std/\allowbreak{}okvqa/\allowbreak{}okvqa & \cite{okvqa} & No license stated (annotations); COCO images & B & 500 \\
img.std/\allowbreak{}scienceqa/\allowbreak{}scienceqa & \cite{lu2022learn} & CC BY-NC-SA 4.0 & A & 500 \\
img.std/\allowbreak{}vizwiz\_qa/\allowbreak{}vizwiz\_vqa & \cite{gurari2018vizwiz} & CC BY 4.0 & B & 500 \\
img.step/\allowbreak{}grounding/\allowbreak{}androidcontrol & \cite{li2024effects} & Apache-2.0 & B & 500 \\
img.step/\allowbreak{}gui\_action/\allowbreak{}mind2web & \cite{zheng2024seeact,deng2023mindweb} & OpenRAIL & B & 249 \\
img.struct/\allowbreak{}chart\_table/\allowbreak{}charxiv & \cite{wang2024charxiv} & CC BY-SA 4.0 (questions); charts by arXiv authors & A & 97 \\
img.struct/\allowbreak{}chart\_table/\allowbreak{}tabmwp & \cite{lu2023dynamic} & CC BY-NC-SA 4.0 & A & 500 \\
img.struct/\allowbreak{}document\_qa/\allowbreak{}infographicvqa & \cite{Mathew_2022_WACV} & annotations CC BY 4.0, images may not be redistributed & B & 500 \\
img.struct/\allowbreak{}form\_fields/\allowbreak{}cord\_v2 & \cite{park2019cord} & CC BY 4.0 & C & 500 \\
\bottomrule
\end{tabular}
\caption{\textbf{Sources of the image tasks.} License as the source states it. Tier of the instructions: A, the source's instructions verbatim; B, adapted from the source; C, newly written from the source's labeling criteria. 3 sources whose annotations we do not redistribute are rebuilt from them by a loader. --: built for this benchmark.}
\label{tab:sources_img}
\end{table*}

\begin{table*}[p]
\centering
\scriptsize
\setlength{\tabcolsep}{3pt}
\begin{tabular}{@{}>{\raggedright\arraybackslash}p{5.2cm}>{\raggedright\arraybackslash}p{1.1cm}%
>{\raggedright\arraybackslash}p{8.6cm}cr@{}}
\toprule
Task and source & Ref. & License & Tier & Questions \\
\midrule
lvid.bench/\allowbreak{}egoschema/\allowbreak{}public\_500 & \cite{mangalam2023egoschema} & Questions: published by the EgoSchema authors, no license stated; clips cut from Ego4D (Ego4D license) & A & 500 \\
lvid.bench/\allowbreak{}vsi\_bench/\allowbreak{}test & \cite{yang2025thinking} & Apache-2.0 (annotations); videos under the ARKitScenes, ScanNet and ScanNet++ terms & A & 500 \\
lvid.dialog/\allowbreak{}meeting\_decision/\allowbreak{}ami & \cite{Carletta_2006} & CC BY 4.0 & B & 500 \\
lvid.embodied/\allowbreak{}ego\_plan/\allowbreak{}egoplan\_bench2 & \cite{qiu2024egoplanbench2} & Questions: published by the EgoPlan-Bench2 authors, no license stated; clips cut from Ego4D (Ego4D license) & A & 500 \\
lvid.judge/\allowbreak{}meeting\_summary/\allowbreak{}tofueval\_meetingbank & \cite{Tang2024,hu-etal-2023-meetingbank} & MeetingBank: CC BY-NC-ND 4.0 (the strictest of its stated licenses); TofuEval: MIT-0 & B & 500 \\
lvid.know/\allowbreak{}lecture/\allowbreak{}videommmu & \cite{hu2025videommmu} & CC BY-NC-SA 4.0 (dataset); videos remain their creators' & A & 500 \\
lvid.know/\allowbreak{}long\_qa/\allowbreak{}video\_mme & \cite{fu2025videomme} & Video-MME terms: academic research only, no distribution without the authors' approval & A & 500 \\
lvid.know/\allowbreak{}needle/\allowbreak{}generated & -- & CC BY 4.0 (generated for this benchmark; no third-party assets) & C & 500 \\
lvid.pro/\allowbreak{}surveillance/\allowbreak{}ucf\_crime & \cite{Sultani_2018_CVPR} & No license stated; video rights not established & C & 500 \\
lvid.ret/\allowbreak{}moment/\allowbreak{}charades\_sta & \cite{Gao_2017_ICCV} & Charades license (non-commercial research) & C & 114 \\
lvid.ret/\allowbreak{}moment/\allowbreak{}ego4d\_nlq & \cite{EGO4D2021} & Ego4D license agreement & C & 247 \\
lvid.ret/\allowbreak{}moment/\allowbreak{}qvhighlights & \cite{NEURIPS2021_62e09734} & CC BY-NC-SA 4.0 (annotations); YouTube videos, no license stated & C & 139 \\
lvid.route/\allowbreak{}escalation/\allowbreak{}jev\_pair & \cite{yang2025thinking,hu2025videommmu,fu2025videomme,Sultani_2018_CVPR,Gao_2017_ICCV,EGO4D2021,NEURIPS2021_62e09734} & Inherits the licenses of the quoted source questions & C & 500 \\
lvid.route/\allowbreak{}segment/\allowbreak{}qvhighlights & \cite{NEURIPS2021_62e09734} & CC BY-NC-SA 4.0 (annotations); YouTube videos, no license stated & B & 500 \\
vid.basic/\allowbreak{}action\_candidates/\allowbreak{}kinetics700 & \cite{smaira2020shortnotekinetics7002020human} & CC BY 4.0 (annotations); YouTube videos keep their own terms & B & 500 \\
vid.basic/\allowbreak{}action\_category/\allowbreak{}ucf101 & \cite{soomro2012ucf101dataset101human} & No license stated (cite Soomro et al., 2012); YouTube videos & B & 500 \\
vid.basic/\allowbreak{}repetition/\allowbreak{}countix & \cite{Dwibedi_2020_CVPR} & No license stated; Kinetics YouTube clips, video rights not established & C & 250 \\
vid.basic/\allowbreak{}repetition/\allowbreak{}repcount\_a & \cite{hu2022transrac} & No license stated; YouTube videos, rights not established & C & 250 \\
vid.basic/\allowbreak{}state\_change/\allowbreak{}changeit & \cite{soucek2022lookforthechange} & MIT (annotations); video rights not established & C & 81 \\
vid.basic/\allowbreak{}state\_change/\allowbreak{}ego4d\_oscc & \cite{EGO4D2021} & Ego4D license agreement & B & 419 \\
vid.basic/\allowbreak{}temporal/\allowbreak{}ssv2 & \cite{Goyal_2017_ICCV,mahdisoltani2018effectivenesstaskgranularitytransfer} & Something-Something v2 license (research use) & C & 500 \\
vid.batch/\allowbreak{}quality/\allowbreak{}konvid\_1k & \cite{konvid1k,hosu2017konstanz} & Free for research, no license stated; Flickr (YFCC100m) videos under their own Creative Commons terms & B & 500 \\
vid.bench/\allowbreak{}mvbench/\allowbreak{}test & \cite{li2024mvbench} & MIT (dataset card); source videos keep their own licenses & A & 500 \\
vid.bench/\allowbreak{}nextqa/\allowbreak{}val & \cite{xiao2021nextqa} & MIT (annotations); VidOR videos from YFCC100m under their own Creative Commons terms & A & 500 \\
vid.bench/\allowbreak{}tempcompass/\allowbreak{}mc & \cite{liu2024tempcompass} & CC BY-NC 4.0 & A & 500 \\
vid.embodied/\allowbreak{}feasibility/\allowbreak{}star & \cite{wu2021star_situated_reasoning} & Charades license (non-commercial research) & A & 500 \\
vid.embodied/\allowbreak{}procedure/\allowbreak{}captaincook4d & \cite{peddi2023captaincook4d} & Apache-2.0 (data); MIT (annotation repository) & B & 500 \\
vid.embodied/\allowbreak{}robot\_success/\allowbreak{}robofac & \cite{ye2026robofaccomprehensiveframeworkrobotic} & MIT & B & 500 \\
vid.fore/\allowbreak{}next\_action/\allowbreak{}epic\_kitchens\_100 & \cite{Damen2022RESCALING,Damen2018EPICKITCHENS} & CC BY-NC 4.0 & B & 500 \\
vid.fore/\allowbreak{}pedestrian/\allowbreak{}jaad & \cite{rasouli2017they,rasouli2018role} & CC BY 4.0 (videos); MIT (annotations) & B & 175 \\
vid.fore/\allowbreak{}pedestrian/\allowbreak{}pie & \cite{Rasouli_2019_ICCV} & MIT (videos and annotations) & B & 325 \\
vid.game/\allowbreak{}physics\_bug/\allowbreak{}physgame & \cite{cao2024physgame} & Apache-2.0 (dataset card); gameplay clips from Reddit and YouTube, rights not established & A & 500 \\
vid.judge/\allowbreak{}hallucination/\allowbreak{}videohallucer & \cite{videohallucer} & MIT (dataset card), CC BY-NC-SA (its paper) & A & 500 \\
vid.judge/\allowbreak{}t2v/\allowbreak{}videofeedback & \cite{he2024videoscore} & Apache-2.0 (dataset card) & B & 500 \\
vid.know/\allowbreak{}causal\_qa/\allowbreak{}perception\_test & \cite{patraucean2023perception} & CC BY 4.0 & A & 500 \\
vid.know/\allowbreak{}physics/\allowbreak{}physion & \cite{bear2021physion} & MIT (code and data) & B & 500 \\
vid.pro/\allowbreak{}driving\_risk/\allowbreak{}nexar & \cite{nexar2025dashcamcollisionprediction} & Nexar Open Data License (attribution; no resale) & B & 500 \\
vid.pro/\allowbreak{}foul/\allowbreak{}soccernet\_mvfoul & \cite{Held_2023_CVPR} & SoccerNet NDA for data and annotations, GPL-3.0 on its Hugging Face card & B & 500 \\
vid.ret/\allowbreak{}text\_match/\allowbreak{}msrvtt\_mc & \cite{Xu_2016_CVPR,Yu_2018_ECCV} & No license stated; video rights not established & B & 500 \\
vid.safety/\allowbreak{}violence/\allowbreak{}ucf\_crime & \cite{Sultani_2018_CVPR} & No license stated; video rights not established & C & 162 \\
vid.safety/\allowbreak{}violence/\allowbreak{}xd\_violence & \cite{Wu2020not} & No license stated; video rights not established & C & 338 \\
\bottomrule
\end{tabular}
\caption{\textbf{Sources of the video tasks.} License and tier as in Table~\ref{tab:sources_img}. 7 sources whose annotations we do not redistribute are rebuilt from them by a loader. --: built for this benchmark.}
\label{tab:sources}
\end{table*}

\begin{table*}[p]
\centering
\scriptsize
\setlength{\tabcolsep}{3pt}
\begin{tabular}{@{}>{\raggedright\arraybackslash}p{5.2cm}>{\raggedright\arraybackslash}p{1.1cm}%
>{\raggedright\arraybackslash}p{8.6cm}cr@{}}
\toprule
Task and source & Ref. & License & Tier & Questions \\
\midrule
aud.basic/\allowbreak{}acoustic\_scene/\allowbreak{}tau2019 & \cite{Mesaros2018_DCASE} & Tampere University terms (non-commercial) & C & 500 \\
aud.basic/\allowbreak{}emotion/\allowbreak{}cremad & \cite{Cao_2014} & ODbL / DbCL; see upstream license & C & 1,470 \\
aud.basic/\allowbreak{}intent/\allowbreak{}slurp & \cite{slurp} & audio CC BY-NC 4.0; annotations CC BY 4.0 & C & 500 \\
aud.basic/\allowbreak{}language\_id/\allowbreak{}voxlingua107\_dev & \cite{valk2021slt} & CC BY 4.0; original copyright retained & C & 1,609 \\
aud.basic/\allowbreak{}sound\_event/\allowbreak{}sonyc\_ust & \cite{Cartwright2020,mark_cartwright_2020_3966543} & CC BY 4.0 & C & 500 \\
aud.basic/\allowbreak{}speaker\_change/\allowbreak{}ami & \cite{Carletta_2006} & CC BY 4.0 & C & 500 \\
aud.dialog/\allowbreak{}dialogue\_state/\allowbreak{}spokenwoz & \cite{spokenwoz_2024} & CC BY-NC 4.0 & C & 500 \\
aud.dialog/\allowbreak{}interrupt\_backchannel/\allowbreak{}full\_duplex\_bench\_v15 & \cite{lin2025fdb_v15} & MIT for synthetic v1.5 data & C & 298 \\
aud.dialog/\allowbreak{}turn\_timing/\allowbreak{}full\_duplex\_bench\_v1 & \cite{lin2025fdb_v1,reece2023candor} & annotations CC BY-NC 4.0, CANDOR audio under the CANDOR Corpus License & C & 335 \\
aud.judge/\allowbreak{}data\_cleaning/\allowbreak{}nisqa\_livetalk & \cite{mittag21_interspeech} & CC BY 4.0; original CC\_BY\_license.txt retained & C & 232 \\
aud.judge/\allowbreak{}data\_cleaning/\allowbreak{}signal\_defects\_synthetic & \cite{slurp} & derived audio CC BY-NC 4.0 & C & 436 \\
aud.judge/\allowbreak{}naturalness/\allowbreak{}bvcc & \cite{cooper21_ssw,huang22f_interspeech} & ratings BSD 3-Clause, audio CC BY 4.0, ODbL, or public domain by source & C & 741 \\
aud.know/\allowbreak{}music\_understanding/\allowbreak{}mmau\_mini & \cite{sakshi2024mmaumassivemultitaskaudio} & CC BY-NC 4.0; underlying source terms apply & C & 105 \\
aud.know/\allowbreak{}reasoning/\allowbreak{}mmau\_mini & \cite{sakshi2024mmaumassivemultitaskaudio} & CC BY-NC 4.0; underlying source terms apply & C & 330 \\
aud.know/\allowbreak{}sound\_qa/\allowbreak{}mmau\_mini & \cite{sakshi2024mmaumassivemultitaskaudio} & CC BY-NC 4.0; underlying source terms apply & C & 36 \\
aud.pro/\allowbreak{}heart\_sound/\allowbreak{}physionet2016 & \cite{Liu_2016,Clifford_2016,pollard_physionet_2026} & Open Data Commons Attribution v1.0; upstream content terms apply & C & 301 \\
aud.pro/\allowbreak{}heart\_sound/\allowbreak{}respiratory\_icbhi2017 & \cite{Rocha_2019} & free for research use with a citation, no license stated & C & 500 \\
aud.pro/\allowbreak{}machine\_anomaly/\allowbreak{}mimii\_pump\_dcase2020 & \cite{Koizumi_DCASE2020_01,Purohit_DCASE2019_01} & CC BY-NC-SA 4.0 & C & 500 \\
aud.ret/\allowbreak{}audio\_text\_matching/\allowbreak{}sonyc\_tag\_descriptions & \cite{Cartwright2020,mark_cartwright_2020_3966543} & CC BY 4.0 & C & 500 \\
aud.safety/\allowbreak{}spoof\_detection/\allowbreak{}asvspoof2019\_la & \cite{todisco19_interspeech} & Open Data Commons Attribution; see original release content conditions & C & 500 \\
aud.safety/\allowbreak{}toxicity/\allowbreak{}mutox & \cite{costajussa2023mutox} & MIT (annotations); audio rights vary & C & 167 \\
\bottomrule
\end{tabular}
\caption{\textbf{Sources of the audio tasks.} License and tier as in Table~\ref{tab:sources_img}.}
\label{tab:sources_aud}
\end{table*}

\section{Closed-Loop Environments}
\label{app:envs}

GameWorld contributes 34 browser games with five tasks each, BALROG 58 tasks, and VS-Bench 10 multi-agent environments
with 10 episodes each.
Each screen goes to the model with its recent history: GameWorld sends the current screenshot and up to two previous
ones, BALROG sends 16 text turns and one image, and VS-Bench sends the images and the observation of the model's role.
The legal actions form one choice question with at most 255 options.
Wordle has more than 255 actions, and the three TextWorld tasks need free-text actions, so a choice question cannot
express them.
NetHack is not played because 100 of its 248 actions cannot be executed through the bridge.
This leaves \nPlayableEnvs{} playable environments.
For Clef and Clef-Flash, the decision prompt of Minesweeper exceeds the input limit, so this environment is not counted
for them.
Table~\ref{tab:loop_success} gives the success rate per suite for the environments that define success.
Table~\ref{tab:realtime} compares real-time and paused GameWorld progress per model.

\begin{table}[t]
\centering
\scriptsize
\setlength{\tabcolsep}{3pt}
\begin{tabular}{@{}lrrrr@{}}
\toprule
Model & GameWorld & BALROG & VS-Bench & Played \\
\midrule
\rowcolor{s1band}\multicolumn{5}{@{}l}{\textit{System-1 models}} \\
d3 & 15.8 & 29.4 & 30.0 & 97/97 \\
Clef & 11.2 & 33.5 & 25.0 & 96/96 \\
OneJev-27B & 18.8 & 21.8 & 25.0 & 97/97 \\
OneJev-9B & 11.5 & 3.8 & 15.0 & 97/97 \\
Clef-Flash & 11.9 & 4.6 & 25.0 & 96/96 \\
OneJev-4B & 11.5 & 3.1 & 27.5 & 97/97 \\
Clef-Omni & 7.9 & 1.7 & 27.5 & 97/97 \\
Xor-26B-A4B & 8.5 & 6.3 & 25.0 & 97/97 \\
OneJev-0.8B & 4.2 & 4.0 & 22.5 & 97/97 \\
Jev-Omni (head v3) & 9.1 & 4.7 & 20.0 & 97/97 \\
Jev-Omni & 8.5 & 5.4 & 22.5 & 97/97 \\
Raya-Decision-v1 & 3.0 & 0.0 & 22.5 & 97/97 \\
\midrule
\rowcolor{s1band}\multicolumn{5}{@{}l}{\textit{Open MLLMs, read at the answer token}} \\
Qwen3.5-35B-A3B (Int4) & 10.9 & 6.9 & 22.5 & 97/97 \\
Qwen3.5-9B & 10.9 & 3.5 & 22.5 & 97/97 \\
Qwen3-VL-30B-A3B & 6.1 & 2.1 & 22.5 & 97/97 \\
Qwen3.8-27B (FP8) & 13.9 & 7.9 & 27.5 & 97/97 \\
Qwen3.5-4B & 13.9 & 3.2 & 22.5 & 97/97 \\
MiniCPM-o 4.5 & 6.7 & 2.3 & 27.5 & 97/97 \\
Qwen3-Omni-30B-A3B (Thinking) & 5.5 & 2.5 & 25.0 & 97/97 \\
Nemotron-3-Nano-Omni & 6.1 & 5.3 & 22.5 & 97/97 \\
Qwen3-Omni-30B-A3B & 9.1 & 2.1 & 42.5 & 97/97 \\
Gemma~4 26B-A4B & 12.7 & 9.7 & 32.5 & 97/97 \\
Qwen2.5-Omni-7B & 7.9 & 3.3 & 27.5 & 97/97 \\
Qwen2.5-Omni-7B (AWQ) & 6.1 & 2.6 & 27.5 & 97/97 \\
Qwen2.5-Omni-7B (GPTQ) & 8.5 & 2.5 & 27.5 & 97/97 \\
DiffusionGemma-26B-A4B (AWQ) & 8.5 & 4.0 & 15.0 & 97/97 \\
DiffusionGemma-26B-A4B & 7.9 & 6.1 & 22.5 & 97/97 \\
Qwen2.5-Omni-3B & 6.7 & 4.4 & 22.5 & 97/97 \\
MiniCPM-o 2.6 & 7.3 & 0.4 & 20.0 & 97/97 \\
Qwen3.5-2B & 4.2 & 1.3 & 22.5 & 97/97 \\
Gemma~4 12B & 9.1 & 7.2 & 50.0 & 97/97 \\
Phi-4-multimodal & 3.6 & 2.3 & 20.0 & 97/97 \\
Qwen3.5-0.8B & 5.5 & 0.0 & 25.0 & 97/97 \\
\bottomrule
\end{tabular}
\caption{\textbf{Closed-loop success rate (\%).} Mean over the environments of a suite that define success, each
environment with equal weight. Played: environments counted, of those the model can take; an environment whose decision
prompt exceeds the model's input limit is not counted.}
\label{tab:loop_success}
\end{table}

\begin{table}[t]
\centering
\scriptsize
\setlength{\tabcolsep}{3pt}
\begin{tabular}{@{}lrrrrr@{}}
\toprule
\multirow{2}{*}{Model} & \multirow{2}{*}{Latency (s)} & \multicolumn{2}{c}{Progress (\%)}
& \multirow{2}{*}{$\Delta$} & \multirow{2}{*}{SE} \\
\cmidrule(lr){3-4}
& & Paused & Real-time & & \\
\midrule
\rowcolor{s1band}\multicolumn{6}{@{}l}{\textit{System-1 models}} \\
d3 & 0.78 & 40.9 & 34.2 & $-$6.7 & 3.8 \\
Clef & 1.14 & 30.9 & 28.1 & $-$2.8 & 4.0 \\
OneJev-27B & 1.09 & 39.8 & 38.2 & $-$1.6 & 3.4 \\
OneJev-9B & 0.45 & 32.2 & 35.7 & +3.4 & 2.7 \\
Clef-Flash & 0.47 & 30.1 & 29.9 & $-$0.3 & 3.2 \\
OneJev-4B & 0.37 & 32.0 & 28.0 & $-$4.0 & 2.3 \\
Clef-Omni & 0.37 & 24.6 & 23.1 & $-$1.5 & 2.3 \\
Xor-26B-A4B & 0.49 & 25.1 & 30.0 & +5.0 & 3.3 \\
OneJev-0.8B & 0.18 & 20.7 & 22.1 & +1.4 & 1.6 \\
Jev-Omni (head v3) & 0.19 & 24.5 & 27.0 & +2.5 & 2.4 \\
Jev-Omni & 0.19 & 25.8 & 24.7 & $-$1.1 & 2.1 \\
Raya-Decision-v1 & 0.23 & 15.2 & 19.9 & +4.8 & 2.7 \\
\midrule
\rowcolor{s1band}\multicolumn{6}{@{}l}{\textit{Open MLLMs, read at the answer token}} \\
Qwen3.5-35B-A3B (Int4) & 0.24 & 31.2 & 33.3 & +2.1 & 2.5 \\
Qwen3.5-9B & 0.24 & 29.5 & 27.5 & $-$2.0 & 3.1 \\
Qwen3-VL-30B-A3B & 0.47 & 23.4 & 27.7 & +4.3 & 2.5 \\
Qwen3.8-27B (FP8) & 0.43 & 34.6 & 33.2 & $-$1.3 & 2.9 \\
Qwen3.5-4B & 0.17 & 31.8 & 25.9 & $-$5.9 & 3.4 \\
MiniCPM-o 4.5 & 0.39 & 21.6 & 21.0 & $-$0.6 & 2.7 \\
Qwen3-Omni-30B-A3B (Thinking) & 0.54 & 20.5 & 26.6 & +6.1 & 2.5 \\
Nemotron-3-Nano-Omni & 0.46 & 18.8 & 22.6 & +3.8 & 2.2 \\
Qwen3-Omni-30B-A3B & 0.53 & 24.6 & 27.3 & +2.7 & 2.4 \\
Gemma~4 26B-A4B & 0.23 & 32.5 & 31.5 & $-$1.0 & 2.4 \\
Qwen2.5-Omni-7B & 0.33 & 23.8 & 22.2 & $-$1.5 & 2.1 \\
Qwen2.5-Omni-7B (AWQ) & 0.34 & 21.2 & 22.0 & +0.8 & 1.5 \\
Qwen2.5-Omni-7B (GPTQ) & 0.33 & 23.4 & 23.4 & 0.0 & 1.5 \\
DiffusionGemma-26B-A4B (AWQ) & 0.24 & 25.6 & 28.2 & +2.6 & 1.9 \\
DiffusionGemma-26B-A4B & 0.23 & 25.9 & 26.2 & +0.3 & 2.0 \\
Qwen2.5-Omni-3B & 0.27 & 17.2 & 20.0 & +2.7 & 2.0 \\
MiniCPM-o 2.6 & 0.43 & 18.3 & 16.8 & $-$1.4 & 0.8 \\
Qwen3.5-2B & 0.14 & 13.6 & 13.9 & +0.3 & 0.4 \\
Gemma~4 12B & 0.17 & 25.1 & 27.5 & +2.4 & 2.1 \\
Phi-4-multimodal & 0.21 & 18.0 & 19.5 & +1.5 & 1.8 \\
Qwen3.5-0.8B & 0.09 & 18.5 & 20.5 & +2.0 & 1.8 \\
\bottomrule
\end{tabular}
\caption{\textbf{GameWorld in real-time mode.} Progress over the games that a model plays in both modes,
33 games with five tasks each (32 for Clef and Clef-Flash). Latency: the game time that real-time mode charges per decision, the
model's median latency on the closed-loop decisions of Table~\ref{tab:latency} in a separate run on H100 PCIe GPUs,
rounded to 10~ms. $\Delta$: real-time minus paused progress (points). SE: the standard error of $\Delta$ over the games.}
\label{tab:realtime}
\end{table}

\section{Spoken Conversations}
\label{app:stream}

\paragraph{Environments.}
The four streaming environments take every recording of four Full-Duplex-Bench sets whose initial audio lasts at most
60~s, \nStreamEpisodes{} episodes in total.
Turn-taking (117 episodes) and pause handling (216 episodes with 277 marked pauses) replay conversations from the
CANDOR Corpus of BetterUp Labs, BetterUp, Inc.\ \cite{reece2023candor}, selected by Full-Duplex-Bench
\cite{lin2025fdb_v1}.
They stay under the CANDOR Corpus License: we release none of their audio, and users obtain it from BetterUp
(\url{https://huggingface.co/datasets/betterup/candor-tabular}).
Interruption (200) and backchannel (98) use the synthetic speech of Full-Duplex-Bench v1.5 \cite{lin2025fdb_v15} under
the MIT license.
Two turn-taking recordings are left out, one of 94~s and one whose held-out turn has zero length.
The audio is converted to 16~kHz mono.
The 80 episodes on which the protocol was tested are part of this pool, so it is not a held-out test set.

\paragraph{Episodes.}
The user audio is sent in packets of 100~ms, each released when its last sample is due, so the conversation goes on
while a model decides.
In turn-taking and pause handling, 5~s of silence follow the user audio.
In interruption and backchannel, the interruption or backchannel starts once the model has spoken audibly for 1~s, and
the episode ends 8~s after it, or after 20~s if the model never speaks that long.
A session lasts at most 110~s.

\paragraph{Decisions.}
A Silero voice activity detector (threshold 0.5, 500~ms of silence, 300~ms of padding) finds the pauses in the user
audio.
At each pause, the model receives a decision request with the user audio observed so far, without its own speech,
transcripts, labels, or future audio.
The state says whether the model is speaking, and the question asks ``Which action should the assistant take now?''
While the model listens, the options are \emph{wait} (``Keep listening because the user has not finished their turn.'')
and \emph{speak} (``Start a response because the user has finished their turn.'').
While it speaks, they are \emph{continue} (``Continue the current response because the user only acknowledges it.'')
and \emph{stop} (``Stop the current response and address the new user request.'').
A hash of the episode and the decision number sets the order of the two options, so a preference for one position does
not favor one action.
Each model returns probabilities as in the offline tasks, and the option with the highest probability is executed.
A decision is discarded if the user speaks again before it returns.

\paragraph{Speech.}
Qwen3-Omni-30B-A3B-Instruct speaks for every model, with its own turn detection off.
After \emph{speak}, it answers the user audio since its last response.
After \emph{stop}, the buffered speech is cleared, the current response is cancelled, and a new response answers the
new user audio.
Speech is played in blocks of 20~ms at 24~kHz and is audible when its RMS exceeds 0.005.

\paragraph{Measures.}
Each episode is scored when its window closes, and it is right-censored if speech is still buffered or generated, or a
decision is pending, at that moment.
In turn-taking, the reference is the moment at which the other speaker took the turn in the recording.
We count the episodes in which the model speaks after it, the median time to its first audible speech, and the episodes
in which it speaks before it.
In pause handling, we count the marked pauses during which the model speaks.
In interruption and backchannel, we count the episodes in which the user utterance was played, and among them the
episodes in which the response is cancelled, which is expected after an interruption and unwanted after a backchannel.
The time to speak includes the 500~ms of silence that the detector waits for, the decision, and the speech generation,
so it is not comparable with Table~\ref{tab:latency}.
The measures are reported separately and not combined into one score (Table~\ref{tab:stream}).

\paragraph{Own speech.}
Two systems also run with their own speech: Qwen3-Omni-30B-A3B-Instruct with the turn policy of its serving runtime,
and MiniCPM-o 4.5 with its own full-duplex mode.
The other models have no validated speech interface in this runtime.
The two rows compare turn policies rather than decisions, so Table~\ref{tab:stream} lists them apart.

\paragraph{Runs.}
All 16 configurations complete the \nStreamEpisodes{} episodes without an execution failure, on RTX A6000 GPUs.
With the shared voice, MiniCPM-o 4.5 and Phi-4-multimodal almost always choose \emph{wait} and never speak in
turn-taking.
Models that rarely speak seldom speak too early or stop for a backchannel, so Table~\ref{tab:stream} gives the counts.

\begin{table}[t]
\centering
\scriptsize
\setlength{\tabcolsep}{3pt}
\begin{adjustbox}{max width=\columnwidth}
\begin{tabular}{@{}lrrrrrrr@{}}
\toprule
\multirow{2}{*}{Model} & \multicolumn{3}{c}{Turn-taking} & Pauses & \multicolumn{2}{c}{Stops after}
& \multirow{2}{*}{Cens.} \\
\cmidrule(lr){2-4} \cmidrule(lr){6-7}
& Spoke & Time (s) & Early & Spoke & Interr. & Backch. & \\
\midrule
\rowcolor{s1band}\multicolumn{8}{@{}l}{\textit{System-1 models}} \\
Clef-Omni & 35 & 1.46 & 1 & 3 & 148/151 & 67/87 & 281 \\
Jev-Omni (head v3) & 65 & 1.32 & 18 & 39 & 99/114 & 35/49 & 353 \\
Jev-Omni & 74 & 1.33 & 17 & 32 & 112/124 & 48/53 & 352 \\
\midrule
\rowcolor{s1band}\multicolumn{8}{@{}l}{\textit{Open MLLMs, read at the answer token}} \\
MiniCPM-o 4.5 & 0 & -- & 0 & 0 & 0/0 & 0/0 & 1 \\
Qwen3-Omni-30B-A3B (Thinking) & 25 & 1.30 & 5 & 5 & 85/86 & 1/1 & 115 \\
Nemotron-3-Nano-Omni & 116 & 1.30 & 32 & 69 & 198/200 & 98/98 & 573 \\
Qwen3-Omni-30B-A3B & 72 & 1.33 & 15 & 17 & 122/125 & 10/59 & 323 \\
Qwen2.5-Omni-7B & 10 & 1.32 & 2 & 4 & 98/101 & 2/11 & 120 \\
Qwen2.5-Omni-7B (AWQ) & 22 & 1.29 & 5 & 11 & 131/142 & 19/48 & 214 \\
Qwen2.5-Omni-7B (GPTQ) & 31 & 1.35 & 5 & 11 & 155/157 & 23/81 & 289 \\
Qwen2.5-Omni-3B & 3 & 1.39 & 0 & 2 & 9/9 & 0/0 & 19 \\
MiniCPM-o 2.6 & 74 & 1.26 & 14 & 22 & 173/179 & 29/58 & 429 \\
Gemma~4 12B & 100 & 1.30 & 35 & 51 & 188/192 & 47/97 & 519 \\
Phi-4-multimodal & 0 & -- & 0 & 0 & 4/4 & 0/0 & 4 \\
\midrule
\rowcolor{s1band}\multicolumn{8}{@{}l}{\textit{Own speech}} \\
MiniCPM-o 4.5, own full-duplex mode & 109 & 1.92 & 11 & 21 & -- & -- & 513 \\
Qwen3-Omni-30B-A3B, runtime turn policy & 117 & 1.15 & 19 & 65 & 199/200 & 98/98 & 594 \\
\bottomrule
\end{tabular}
\end{adjustbox}
\caption{\textbf{Spoken conversations.} Counts over the \nStreamEpisodes{} episodes. Turn-taking
(117 episodes): Spoke counts the episodes in which the model speaks after the
moment at which the other speaker took the turn in the recording, Time is the median time from that moment to its first
audible speech (0 if it is already speaking then), and Early counts the episodes in which it speaks before that moment.
Pauses: the marked pauses, of 277, during which it speaks.
Stops after: cancelled responses out of the episodes in which the interruption
(of 200) or the backchannel (of 98) was played.
Stopping is expected after an interruption and unwanted after a backchannel. Cens.: right-censored episodes. With the
shared voice, Qwen3-Omni-30B-A3B-Instruct speaks for every model, and Time includes the 500~ms of silence that the
voice activity detector waits for. Own speech: the system speaks with its own voice and turn policy, so these rows do
not compare decisions, and MiniCPM-o 4.5 in its own mode reports no cancellation, so its stops are not counted.}
\label{tab:stream}
\end{table}

\section{Limitations}
\label{app:limitations}
Every label comes from its source, so the source's label noise carries over, and many sources are public, so some videos
may have been seen in training.
Real-time mode charges every decision the model's median latency on H100 GPUs at batch size 1, and slower hardware,
shared servers, or single slow decisions could cost progress that these results do not show.
In the streaming environments, every model speaks through one shared speech model and decides only at detected pauses.
The runs with thinking sample one thought per question and stop it at 2{,}048 tokens, which up to 39\% of the thoughts
on video and 92\% of the decisions in closed loops reach.

\section{Ethical Considerations}
\label{app:ethics}
\name{} reuses public datasets and environments under their licenses (Section~\ref{app:sources}).
We host the media that their licenses let us redistribute in a gated dataset, together with the license texts and
per-file credits those licenses ask for.
Media whose terms forbid redistribution, or allow sharing only with people who accept them, are downloaded by each user
from the publisher, with the user's own access where the publisher asks for it.
For the \nRecipeSourcesRelease{} sources whose annotations we do not redistribute, we release only how each question is
built, and a loader rebuilds the questions locally from the source's own release.
Surveillance and violence tasks contain disturbing content by design.
We keep these videos as their creators released them.
A high score does not make a model safe to trust with decisions that have legal, safety, or medical consequences.
In such decisions, an error can harm people, and a person should still review the outcome.

\end{document}